\documentclass[review,authoryear]{elsarticle}

\usepackage{lineno,hyperref}
\usepackage{amsmath}
\allowdisplaybreaks
\usepackage{breqn}
\usepackage{calc}  
\usepackage{enumitem}  
\usepackage{amsfonts}
\usepackage{mathrsfs}
\usepackage{xcolor}
\usepackage{natbib}
\usepackage{bigstrut}
\usepackage{booktabs}
\usepackage{rotating}
\usepackage{tabularx}
\usepackage{hyperref}
\usepackage[ruled,vlined]{algorithm2e} 

\usepackage[top=3.5cm,bottom=3.5cm,left=3.5cm,right=3.5cm]{geometry}
\usepackage{caption}
\usepackage{subcaption}
\usepackage{tikz}
\modulolinenumbers[5]
\usepackage{graphicx, amsfonts,amsmath , boldline, epstopdf,amssymb }
\usepackage{tabu, multirow,multicol ,booktabs ,setspace ,algcompatible ,mathrsfs,dsfont, array, boldline, makecell, booktabs }
\usepackage{url}
\usepackage{epstopdf}
\usepackage{adjustbox, stackengine, color, colortbl, epsfig, cases, enumitem, mathtools, tikz, verbatim}
\usetikzlibrary{shapes,arrows.meta,decorations}
\definecolor{Abi}{rgb}{0.309803, 0.58039, 0.80392}
\definecolor{orange}{rgb}{1, 0.5019, 0}
\definecolor{Red}{rgb}{1, 0, 0}
\definecolor{orange}{rgb}{1, 0.50196, 0}
\definecolor{greenJ}{rgb}{0, 0.6590, 0.42}
\definecolor{Brown}{rgb}{0.588, 0.294, 0}
\definecolor{um}{rgb}{0.0824, 0.1294, 0.4196}
\definecolor{abikam}{rgb}{0.51, 0.93,  0.992}
\usepackage{amsthm}

\usepackage{lineno,hyperref}
\usepackage{amsmath}
\usepackage{breqn}
\usepackage{calc}  
\usepackage{enumitem}  
\usepackage{amsfonts}
\usepackage{mathrsfs}
\usepackage{multirow}
\usepackage{bigstrut}
\usepackage{booktabs}
\usepackage{rotating}
\usepackage{hyperref}
\usepackage{nomencl}

\makenomenclature
\renewcommand\nomgroup[1]{%
  \item[\bfseries
  \ifstrequal{#1}{I}{Sets and Indices}{%
  \ifstrequal{#1}{P}{Parameters}{%
  \ifstrequal{#1}{L}{Functions}{%
  \ifstrequal{#1}{V}{Variables}{}}}}%
]}

\usepackage{bm}
\usepackage[top=3.5cm,bottom=3.5cm,left=3.5cm,right=3.5cm]{geometry}
\usepackage{caption}
 
\modulolinenumbers[5]
\usepackage{mathtools}
\usepackage{stackengine}

\usepackage{lineno,hyperref}
\usepackage{amsmath}
\usepackage{breqn}
\usepackage{calc}  
\usepackage{enumitem}  
\usepackage{amsfonts}
\usepackage{mathrsfs}
\usepackage{multirow}
\usepackage{bigstrut}
\usepackage{booktabs}
\usepackage{rotating}
\usepackage{hyperref}
\usepackage[top=3.5cm,bottom=3.5cm,left=3.5cm,right=3.5cm]{geometry}

\usepackage{xcolor}
\usepackage{soul}
\usepackage{comment}

\usepackage{amsmath, amssymb}

 \usepackage{caption}

\usepackage{color}

\newcommand{\reals}{{I\kern-.35em R}}
\newcommand{\nats}{{I\kern-.35em N}}
\newcommand{\upto}{{\raise 1pt \hbox{$\scriptstyle \,\nearrow\,$}}}
\newcommand{\downto}{{\raise 1pt \hbox{$\scriptstyle \,\searrow\,$}}}
\newcommand{\eop}
	{\hfill{$\vcenter{\hrule height1pt \hbox{\vrule width1pt height5pt 
   	 \kern5pt \vrule width1pt} \hrule height1pt}$} \medskip}
\def\state #1. { \noindent{\bf#1.\enspace}}

\usepackage{lineno,hyperref}
\usepackage{amsmath}
\usepackage{breqn}
\usepackage{calc}  
\usepackage{enumitem}  
\usepackage{amsfonts}
\usepackage{mathrsfs}

\usepackage{xcolor}
\usepackage{bigstrut}
\usepackage{rotating}
\usepackage[top=3.5cm,bottom=3.5cm,left=3.5cm,right=3.5cm]{geometry}

\usepackage{subcaption}
\modulolinenumbers[5]
\usepackage{graphicx, amsfonts,amsmath , boldline, epstopdf,amssymb }
\usepackage{tabu, multirow,multicol ,booktabs ,setspace ,algcompatible ,mathrsfs,dsfont, array, boldline, makecell, booktabs }
\usepackage{url}
\usepackage{epstopdf}
\usepackage{adjustbox, stackengine, color, colortbl, epsfig, cases, enumitem, mathtools, tikz, verbatim}
\usetikzlibrary{shapes,arrows.meta,decorations}
\definecolor{um}{rgb}{0.0824, 0.1294, 0.4196}
\definecolor{Abi}{rgb}{0.059, 0.322, 0.729}
\definecolor{orange}{rgb}{1, 0.347, 0}
\definecolor{c}{rgb}{0, 1, 1}
\definecolor{m}{rgb}{1, 0, 1}

\definecolor{bb}{rgb}{0.2941, 0.5447, 0.7494}
\definecolor{greenJ}{rgb}{0, 0.6590, 0.42}
\definecolor{zereshk}{rgb}{0.588,0,0.098}
\definecolor{Maroon}{rgb}{0.502, 0, 0}
\definecolor{Brown}{rgb}{0.588, 0.294, 0}
\definecolor{Olive}{rgb}{0.502, 0.502, 0}
\definecolor{Navy}{rgb}{0, 0, 0.502}
\definecolor{Orange}{rgb}{1,0.647,0}
\definecolor{Yellow}{rgb}{0.502, 1, 0}
\definecolor{Green}{rgb}{0, 0.502,  0}
\definecolor{Blue}{rgb}{0, 0,0.761}
\definecolor{Lime}{rgb}{0.196, 0.804, 0.196}
\definecolor{Purple}{rgb}{0.502,0,0.502}
\definecolor{Violet}{rgb}{0.561,0,1}
\definecolor{Magneta}{rgb}{1,0,1}
\definecolor{Red}{rgb}{1,0,0}
\definecolor{Gray}{rgb}{0.502, 0.502, 0.502}
\definecolor{abikam}{rgb}{0.51, 0.93,  0.992}
\definecolor{abizeyad}{rgb}{0.16, 0.573,0.761}
\definecolor{rr}{rgb}{0.9047, 0.1918, 0.1988}

\definecolor{orange2}{rgb}{0.85,0.33,0.10}
\definecolor{yellow4}{rgb}{0.93,0.69,0.13}
\definecolor{purple6}{rgb}{0.49,0.18,0.56}
\definecolor{green32}{rgb}{0.47,0.67,0.19}
\definecolor{blue13}{rgb}{0.30,0.75,0.93}
\definecolor{zereshk16}{rgb}{0.64,0.08,0.18}
\definecolor{orange25}{rgb}{0.93,0.69,0.13}
\definecolor{gray27}{rgb}{0.0824, 0.1294, 0.4196}

\usetikzlibrary{tikzmark}

\tikzset{mycircled/.style={circle,draw,inner sep=0.1em,line width=0.04em}}

\tikzset{mycircled2/.style={circle,draw,inner sep=0.1em,line width=0.09em}}

 \usepackage{caption}
 
\journal{Engineering Applications of Artificial Intelligence}

\begin{document}
\doublespacing
\begin{frontmatter}

\title{Reinforcement Learning-Guided Dynamic Multi-Graph Fusion for Evacuation Traffic Prediction}




\author{Md Nafees Fuad Rafi\corref{mycorrespondingauthor}}
\author{Samiul Hasan}

\address{Department of Civil, Environmental and Construction Engineering, University of Central Florida, 12800 Pegasus Drive, Orlando, FL 32816, United States}

\cortext[mycorrespondingauthor]{Corresponding author: mdnafeesfuad.rafi@ucf.edu}


\begin{abstract}
Real-time traffic prediction is critical for managing transportation systems during hurricane evacuations. Although data-driven graph-learning models have demonstrated strong capabilities in capturing the complex spatiotemporal dynamics of evacuation traffic at a network level, they mostly consider a single dimension (e.g., travel-time or distance) to construct the underlying graph. Furthermore, these models often lack interpretability, offering little insight into which input variables contribute most to their predictive performance. To overcome these limitations, we develop a novel Reinforcement Learning-guided Dynamic Multi-Graph Fusion (RL-DMF) framework for evacuation traffic prediction. We construct multiple dynamic graphs at each time step to represent heterogeneous spatiotemporal relationships between traffic detectors. A dynamic multi-graph fusion (DMF) module is employed to adaptively learn and combine information from these graphs. To enhance model interpretability, we introduce RL-based intelligent feature selection and ranking (RL-IFSR) method that learns to mask irrelevant features during model training. The model is evaluated using a real-world dataset of 12 hurricanes affecting Florida from 2016 to 2024. For an unseen hurricane (Milton, 2024), the model achieves a 95\% accuracy (RMSE = 293.9) for predicting the next 1-hour traffic flow. Moreover, the model can forecast traffic flow for up to next 6 hours with 90\% accuracy (RMSE = 426.4). The RL-DMF framework outperforms several state-of-the-art traffic prediction models. Furthermore, ablation experiments confirm the effectiveness of dynamic multi-graph fusion and RL-IFSR approaches for improving model performance. This research provides a generalized and interpretable model for real-time evacuation traffic forecasting, with significant implications for evacuation traffic management.
\end{abstract}

\begin{keyword}
Dynamic Multi-Graph Fusion, Reinforcement Learning, Double Deep Q-Network, Hurricane Evacuation, Traffic Networks, Traffic Prediction
\end{keyword}
\end{frontmatter}

\section{Introduction}
Real-time traffic prediction has potential for better managing transportation systems during hurricane evacuations. Effective evacuation planning and traffic management can reduce travel time and the number of crashes by enabling pro-active traffic management and routing strategies \cite{RAHMAN2023104126, jiang2024scalable_planning}. However, prediction models developed for regular traffic conditions often fail during evacuation due to sudden surges and abrupt evacuation dynamics. Traffic prediction for an evacuation period has several challenges: (i) the spatiotemporal patterns of traffic congestion often change rapidly over a short horizon, leading to distribution shift in the data \cite{RAHMAN2023104126, jiang2024scalable_planning}; (ii) data scarcity during evacuations complicates model training, since historical high-resolution datasets are limited \cite{rashid2024evacuation}; and (iii) human behavior such as evacuation destination, departure times, and route choices are heterogeneous and unpredictable \cite{jiang2024scalable_planning}. These challenges warrant models that can adapt in real time to rapidly evolving evacuation traffic conditions.

Data-driven deep learning models have emerged as powerful tools for traffic forecasting due to their ability to capture complex spatiotemporal relationships \cite{jiang2021gnn_survey, zhang2025dynamic_traffic}. Combining deep learning architectures such as Graph Neural Network (GNN), Convolutional Neural Network (CNN), Long-short Term Memory (LSTM) models can handle high dimensional spatial-temporal traffic data very effectively, thus making more accurate predictions than traditional machine learning or statistical approaches \cite{RAHMAN2023104126}. As transportation networks can be represented as a graph, GNN models generally perform well in capturing the spatio-temporal variation of evacuation traffic states. Particularly, graph convolutional network (GCN) architecture \cite{kipf2017semi} has been recently adopted due to its advantages in dealing with non-Euclidean and irregular data, such as transportation network data \cite{chen2022reinforced}. For instance, previous studies \cite{rashid2024evacuation} and \cite{RAHMAN2023104126} developed Graph Convolutional Long-Short Term Memory (GCN-LSTM) models to predict traffic in major highways during hurricane evacuations. These studies showed that graph-based models can capture evacuation traffic better than other traditional machine learning or deep learning approaches.

Another limitation of traditional GNN models is that it considers the graph-based transportation network to be static as the number of nodes remain same over time. However, traffic detectors (i.e., graph nodes) may occasionally go offline due to scheduled maintenance, power outage or other technical reasons. Although current graph-based evacuation traffic prediction models \cite{rashid2024evacuation, RAHMAN2023104126} capture the dynamic variation in traffic features, they fail to capture the dynamics of the graph topology itself. Furthermore, detector-based network topology varies across multiple hurricanes as existing detectors may go out of order permanently and new detectors may be installed to cover unobserved locations \cite{rafi2025dynamic}. To overcome this limitation, \cite{rafi2025dynamic} proposed a dynamic graph learning framework to capture network dynamics for predicting traffic during hurricane evacuation. However, they considered distance as the edge weight of dynamic graph. In their study, although the network topology is changing with time, the weights (distance between a node pair) of a given edge of a dynamic graph does not change with time. Moreover, distance only represents the graph in a static manner, independent of congestion or speed. Hence, a graph learning model trained using only distance as edge weight could miss important information during evacuation. Travel time encodes dynamic traffic conditions more accurately as it reflects real-time conditions due to congestion and incidents. But travel time alone might lack context about spatial layout of the dynamic graph. Using both distance and travel time as edge weights of the dynamic graph may make the model more sensitive to real-time traffic conditions and better reflect actual congestion dynamics across the network during hurricane evacuation. \cite{li2023dmgfnet} adopted a graph fusion technique to model the interaction of multiple spatial correlation to improve prediction accuracy. However, they built their model considering regular period traffic, which is significantly different than evacuation traffic. Moreover, they did not consider the evolving structure of the dynamic graph. To the best of our knowledge, no existing study simultaneously model multiple spatial dependencies (distance and travel time) together for capturing the real-time congestion dynamics of evacuation traffic flow.

Although deep learning-based traffic prediction models have high accuracy, it is hard to interpret which input features contribute more to prediction performance \cite{burrell2016blackbox, adadi2018peeking}. This lack of interpretability may hinder the real-world deployment of models, especially under non-recurrent conditions like hurricane evacuations since decision-makers cannot explain any unexpected model predictions and adjust strategies in response to such predictions when the influence of each feature remains unclear. Traditional feature selection methods such as filter-based or embedded techniques typically operate independently of model performance or ignore spatio-temporal dependencies present in traffic dynamics. To overcome this limitation, reinforcement learning (RL) offers a promising alternative by framing feature selection as a sequential decision-making process, where an agent learns to select informative features based on their cumulative contribution to prediction performance. Recent studies have shown that RL-based feature selection methods outperform static selection techniques in domains like biomedical data \cite{du2025reinforcement} and financial forecasting \cite{bai2024rlfinance}. However, the potential of this approach for evacuation traffic prediction tasks under dynamic network conditions is yet to be explored. 

To address these gaps, we develop a novel Reinforcement learning-guided Dynamic Multi-graph Fusion (RL-DMF) architecture for evacuation traffic prediction. It integrates: (i) an attention-based fusion across multiple dynamic graphs; and (ii) an RL agent that learns to select features adaptively through intelligent feature masking. First, the proposed model constructs two dynamic graphs at each time step to represent physical and travel‐time relationships. Then, these graphs are fused together by adopting an attention mechanism, where both graph's output are weighted with a learnable attention score. The fused outputs are passed through an LSTM layer and finally, a linear layer to predict the evacuation traffic flow. We train an RL-agent simultaneously which learns to mask irrelevant feature during each training step. Specifically, we introduce a Double Deep Q-network \cite{hasselt2010double} agent that learn to emphasize important feature during model training and improves the model robustness and generalizability—especially in dynamic and data-scarce environments like evacuation periods. The RL-agent gives a feature ranking based on the masking frequency of individual features of the model. 

To implement this framework, we predict evacuation traffic in five major highways of Florida. We used traffic data from twelve (12) historical hurricanes which made landfall in Florida from 2016 to 2024. For each hurricane, we collected ten (10) consecutive days of traffic data including non-evacuation period, evacuation period, and landfall day data. We also collected evacuation and incident related data during each of the hurricane and integrated it with traffic data. We used data from past 6 hours to predict future traffic flow for next 6 hours. We tested the model on the two-day evacuation period of Hurricane Milton: October 7 and 8, 2024. To test the generalization capability of the model, we also trained another model on past hurricanes from Hurricane Hermine (2016) to Hurricane Elsa (2021) and tested on Hurricane Ian (2022). 

This framework gives us a dynamic and generalized evacuation traffic prediction model and an intelligent feature ranking process, enabling an interpretable model for any future hurricane in Florida. The contributions of this study are summarized as follows:

\begin{enumerate}
    \item We develop an attention-based dynamic multi-graph fusion (DMF) module that fuses both distance and travel time–based dynamic graphs to capture heterogeneous spatio-temporal dependencies and evolving traffic conditions during evacuations.
    
    \item We introduce an RL-based Intelligent Feature Selection and Ranking (RL-IFSR) method that learns to perform intelligent, context-aware feature masking, improving model interpretability and generalizability.

    \item Integrating DMF and RL-IFSR, we develop a novel RL-Guided Dynamic Multi-Graph Fusion (RL-DMF) architecture for network-wide evacuation traffic prediction of any future hurricane in Florida.
    
    \item We demonstrate that DMF improves prediction performance over using either graph alone and the RL-based feature masking leads to more robust and interpretable predictions under dynamic and data-scarce evacuation conditions. 

    \item We evaluate the model's generalizability for different hurricanes and different parts of the network--an aspect that is largely overlooked in existing literature.

\end{enumerate}

\section{Study Area and Data Preparation} \label{sec:data}
\begin{figure}[htbp]
    \centering
    \begin{subfigure}[b]{0.45\linewidth}
        \centering
        \includegraphics[width=\linewidth]{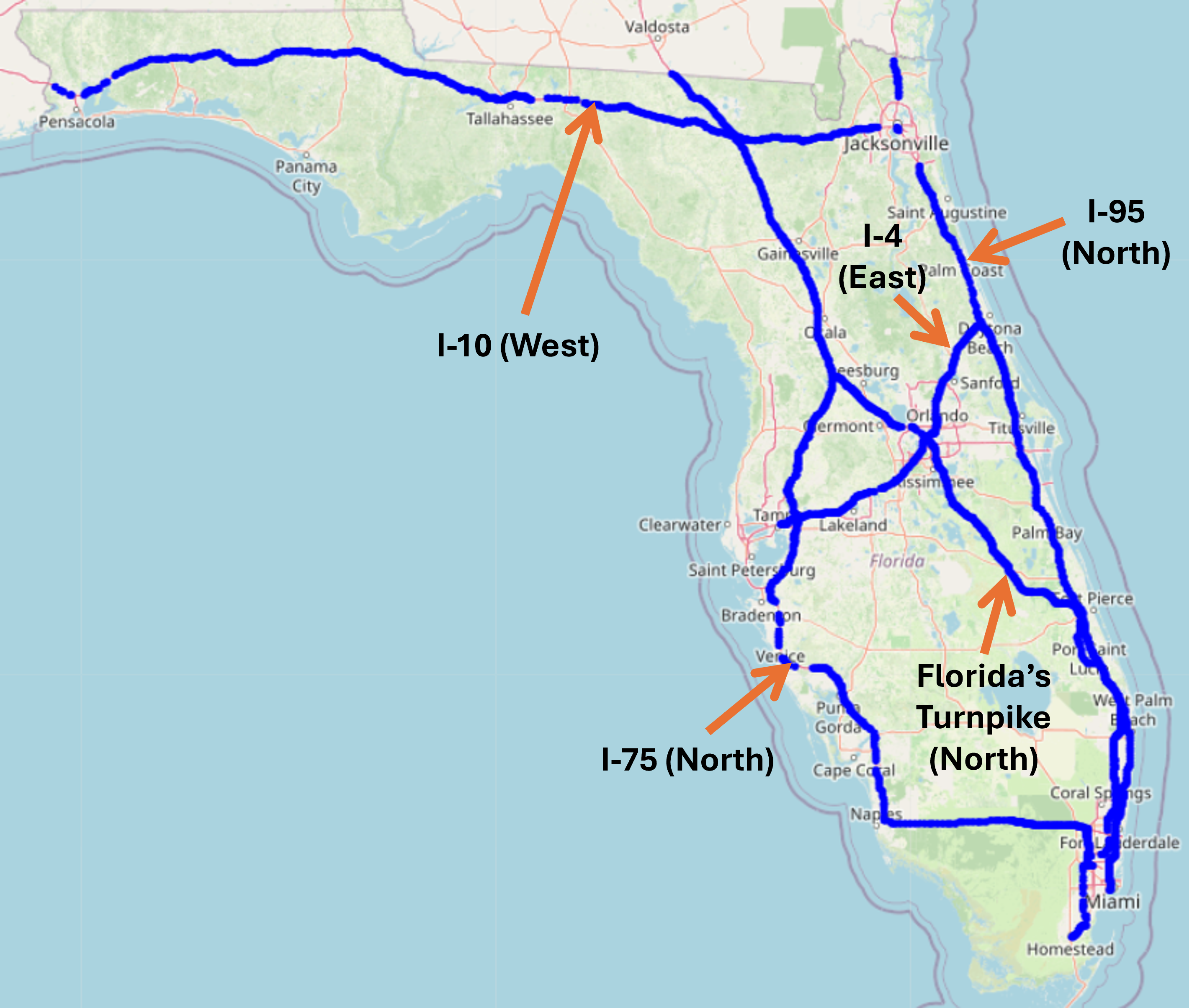}
        \caption{Florida traffic detector distribution in highways}
        \label{fig:det_plot}
    \end{subfigure}
    \hfill
    \begin{subfigure}[b]{0.50\linewidth}
        \centering
        \includegraphics[width=\linewidth]{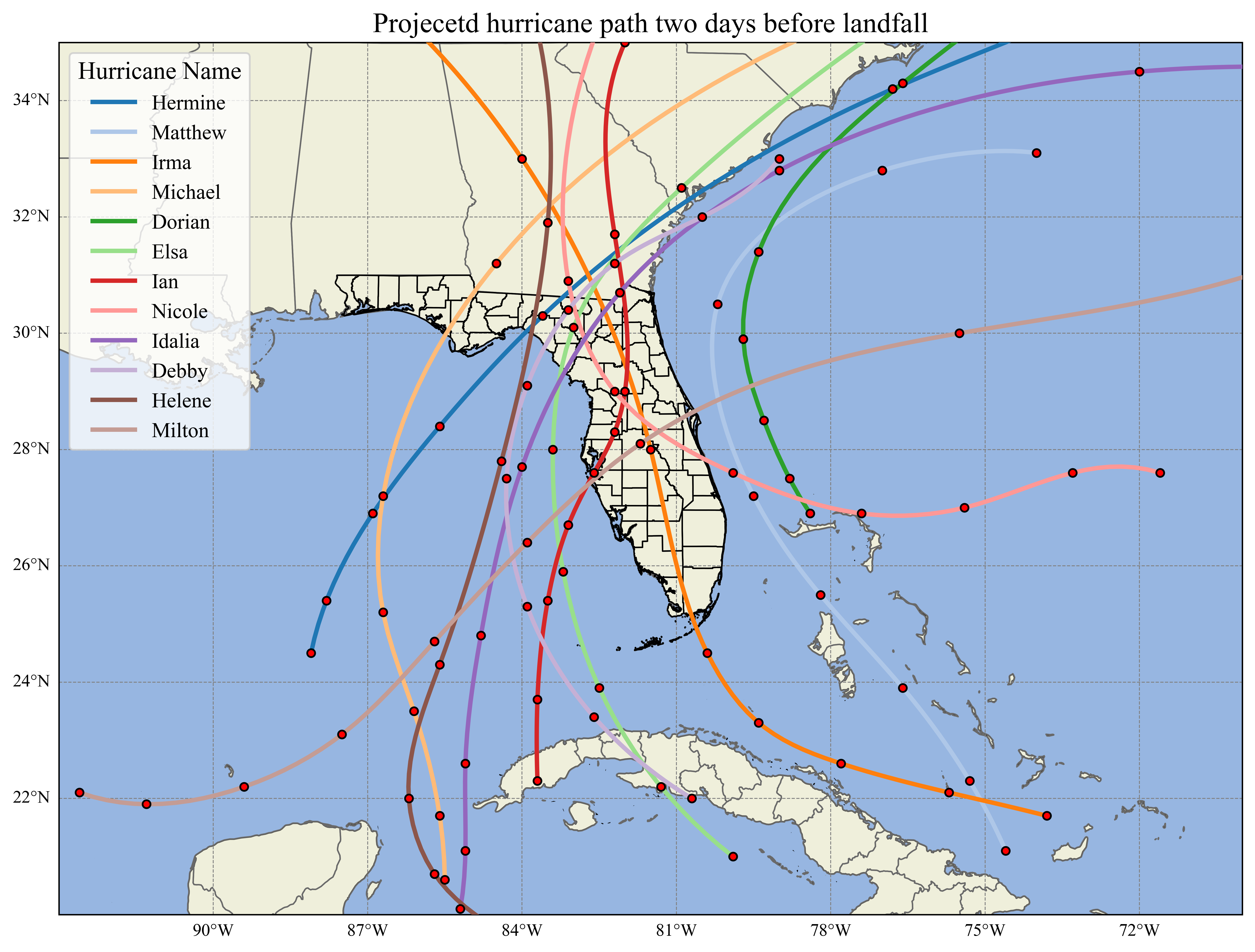}
        \caption{Projected hurricane paths 2 days before landfall}
        \label{fig:hurricane_path}
    \end{subfigure}
    \caption{Traffic detectors and hurricane paths in Florida}
    \label{fig:detectors_and_hurricane_path}
\end{figure}

In this study, we consider a transportation network comprising five major interstate highways in Florida: I-4 (East), I-10 (West), I-75 (North), I-95 (North), and Florida's Turnpike (North). As illustrated in Figure~\ref{fig:det_plot}, these highways are equipped with traffic detectors and serve as primary evacuation corridors during hurricanes, facilitating population movement toward Georgia and other neighboring states \cite{RAHMAN2023104126}. We obtained traffic data for these corridors from the Regional Integrated Transportation Information System (RITIS) \cite{ritis_archive}, which aggregates high-resolution detector-level volume, speed, and occupancy data collected via the Microwave Vehicle Detection System (MVDS) maintained by the Florida Department of Transportation (FDOT). 

For each hurricane event, we collected hourly traffic data over a 10-day period (see Table~\ref{tab:hurricane_details}), covering non-evacuation, evacuation, and landfall phases. Raw traffic detector data often suffer from errors caused by sensor malfunctions, extreme weather, and missing or duplicate entries, especially under evacuation-induced congestion conditions where MVDS detectors can fail to register vehicles \cite{RAHMAN2023104126, rashid2024evacuation}. To ensure data reliability, we processed the raw traffic data by adopting the data cleaning pipeline proposed by \cite{RAHMAN2023104126}.

\begin{table}[h]
\centering
\caption{Hurricane Information and RITIS Data Collection Period}
\label{tab:hurricane_details}
\begin{tabularx}{\linewidth}{|p{3.5cm}|p{5cm}|X|}
\hline
\textbf{Hurricane Name} & \textbf{Landfall Date} & \textbf{Data Range} \\
\hline
Hermine & September 2, 2016 & 08/24/2016 – 09/02/2016 \\
Matthew & October 7, 2016 & 09/28/2016 – 10/07/2016 \\
Irma & September 10, 2017 & 09/01/2017 – 09/10/2017 \\
Michael & October 10, 2018 & 10/01/2018 – 10/10/2018 \\
Dorian & September 4, 2019 & 08/26/2019 – 09/04/2019 \\
Elsa & July 7, 2021 & 06/28/2021 – 07/07/2021 \\
Ian & September 28, 2022 & 09/19/2022 – 09/28/2022 \\
Nicole & November 10, 2022 & 11/01/2022 – 11/10/2022 \\
Idalia & August 30, 2023 & 08/21/2023 – 08/30/2023 \\
Debby & August 5, 2024 & 07/27/2024 – 08/05/2024 \\
Helene & September 27, 2024 & 09/18/2024 – 09/27/2024 \\
Milton & October 10, 2024 & 10/01/2024 – 10/10/2024 \\
\hline
\end{tabularx}
\end{table}

Hurricane paths are inherently dynamic and uncertain, making the evacuation process a complex task. Figure~\ref{fig:hurricane_path} illustrates the projected paths of the historical hurricanes, predicted two days prior to their landfall days. These projections significantly influence evacuation decisions as emergency management authorities rely on them to issue timely evacuation orders across different regions. For most hurricanes, evacuation order data were obtained from official sources such as \cite{florida_evac_orders} and \cite{Anand2024HEvOD}. However, for several recent hurricanes in 2024, such information was unavailable through conventional sources. In those cases, we retrieved evacuation order announcements from the official social media platforms of the respective counties (\cite{facebook_homepage}, \cite{x_homepage}), where real-time updates were posted. Evacuation-related information was compiled for each hurricane listed in Table~\ref{tab:hurricane_details}. To better understand the scale of evacuation, we also collected population data for the evacuation zones for each of the hurricanes.

A list of all features considered in this study is given in Table~\ref{tab:features}. We extracted typical traffic and evacuation related features similar to \cite{RAHMAN2023104126, rashid2024evacuation} along with some new features such as weekday, lane numbers, landfall day, evacuation day, distance from landfall location and time elapsed after evacuation order. Additionally, to capture the impact of incidents on evacuation traffic flow, we collected incident data from \cite{ritis_incident_archive}. We extracted incident-related features using KDTree-based spatial matching \cite{panigrahy2008improved}, associating each detector-time pair with nearby incident records. These include binary incident flags, number of lanes closed, total number of incidents, incident duration statistics, and time since the last incident, which may help the model learn the impact of disruptions on evacuation traffic. 

\begin{table}[htbp]
\centering
\caption{Model Input Features}
\label{tab:features}
\footnotesize 
\begin{tabularx}{\linewidth}{XXX}
\toprule
\textbf{Traffic Features} & \textbf{Incident Features} & \textbf{Evacuation Features} \\
\midrule
Detector ID & Incident & Cum. pop. under evc. orders \\
Time periods & Number of incidents & Distance from nearest evc. zones \\
Traffic flow & Maximum lane closed & Distance from landfall location \\
Traffic speed & Total number of vehicles involved & Time left before landfall \\
Previous day mean traffic flow & Avg. incident duration & Time elapsed after evc. order \\
Previous period mean traffic flow & Max. incident duration & Evacuation day \\
Previous day std. of traffic flow & Avg. time elapsed after incident & Landfall day \\
Previous period std. of traffic flow & Max. time elapsed after incident & \\
Highway (I-4, I-10, I-75, I-95, TP)&  & \\
Weekday &  & \\
Lane Numbers &  & \\
\bottomrule
\end{tabularx}
\end{table}

\section{Methodology} \label{sec:methodology}
\subsection{DMF: Dynamic Multi-Graph Fusion} \label{sec:dmf}
\begin{figure}[htbp]
    \centering
    \includegraphics[width=0.95\linewidth]{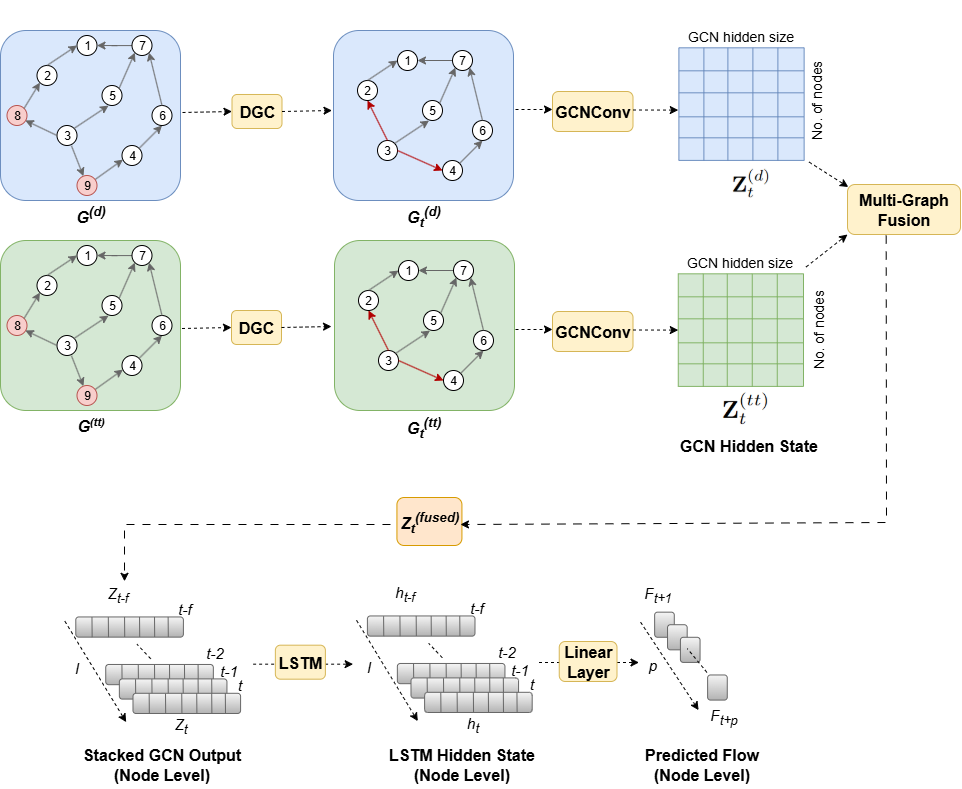}
    \caption{Dynamic Multi-Graph Fusion (DMF) Framework}
    \label{fig:dmf}
\end{figure}

We consider a detector-based transportation network graph with five major highways in Florida, where each detector is modeled as a node and the road segment connecting adjacent detectors is represented as an edge. To capture spatial-temporal dependencies, we develop a dynamic multi-graph fusion architecture (see Figure~\ref{fig:dmf}). This module integrates information from two graph types: distance-based and travel-time-based graphs. The distance-based graph captures static physical distance between detectors and the travel-time-based graph reflects dynamic congestion and delays that evolve during evacuation. By incorporating these two graphs, the model learns to capture both long-term structural and short-term operational dynamics.

Let \( \tilde{\mathbf{X}}_{\text{temp}} \in \mathbb{R}^{N_t \times T \times F_t} \) denote the time-varying temporal feature matrix over \( l \) historical steps and \( F_t \) denotes the vector of temporal features, where \( N_t \) is the number of active detectors (nodes) at time step \( t \). Let \( \tilde{\mathbf{X}}_{\text{spatial}} \in \mathbb{R}^{N_t \times F_s} \) denote the static spatial feature matrix, with \( F_s \) features per node. At each time step \( t \in \{1, \dots, l\} \), the temporal and spatial features are concatenated to construct a unified node representation:

\begin{equation} \label{eq:input_features}
\mathbf{H}_t = \tilde{\mathbf{X}}_{\text{temp}}[:, t, :] \, \Vert \, \tilde{\mathbf{X}}_{\text{spatial}} \in \mathbb{R}^{N_t \times (F_t + F_s)}, \quad \forall t \in l
\end{equation}

The dynamic graph construction is a key step in enabling the model to adapt to real-time evacuation traffic conditions. In this work, we build upon a dynamic graph construction (DGC) framework developed in \cite{rafi2025dynamic} which accounts for dynamic detector topology across time. At each time step \( t \), we construct two separate graphs over the node set \( V_t \), each capturing different aspects of network connectivity. These graphs are defined as follows:
\begin{equation}
G_t^d = (V_t, E_t^d, \mathbf{A}_{t}^{d})
\end{equation}
\begin{equation}
G_t^{tt} = (V_t, E_t^{tt}, \mathbf{A}_{t}^{tt})
\end{equation}
Here, \( G_t^d \) denotes the distance-based graph, where \( \mathbf{A}_{t}^d \) is the distance-based adjacency matrix at time \( t \). \( G_t^{tt} \) denotes the travel-time-based graph with corresponding adjacency matrix \( \mathbf{A}_{t}^{tt} \). \( E_t^d \) and \( E_t^{tt} \) denotes the edge sets of the distance and travel time based graph at $t$, respectively. For the travel-time-based graph, we determine the dynamic travel time ($tt_t$) between detectors as follows.

\begin{equation}
tt_t(i, j) = \frac{d(i, j)}{\frac{v_t(i) + v_t(j)}{2}}
\label{eq:travel_time}
\end{equation}

where \( v_t(i) \) and \( v_t(j) \) represent the speed at detectors \( i \) and \( j \) at time \( t \), respectively. \( d(i, j) \) denotes the distance between nodes \( i \) and \( j \).

The adjacency matrices are defined as follows.

For the distance-based graph,
\begin{equation}
\mathbf{A}_t^{d}[i, j] = 
\begin{cases} 
d(i, j) & \text{if } (i, j) \in E_t^d \\ 
0 & \text{otherwise}
\end{cases}
\label{eq:A_t_distance}
\end{equation}

For the travel-time-based graph,
\begin{equation}
\mathbf{A}_t^{tt}[i, j] = 
\begin{cases} 
tt_t(i, j) & \text{if } (i, j) \in E_t^{tt} \\ 
0 & \text{otherwise}
\end{cases}
\label{eq:A_t_traveltime}
\end{equation}
This formulation allows the travel-time-based graph to dynamically reflect real-time traffic conditions. The edge weights in both \( \mathbf{A}_{t}^{d} \) and \( \mathbf{A}_{t}^{tt} \) are normalized using min-max scaling to maintain consistency in magnitude.

To capture complementary spatial relationships from different graph modalities, we use two separate graph convolutional networks (GCNs), each dedicated to a specific graph structure: the distance-based graph \( G_t^d \) and the travel-time-based graph \( G_t^{tt} \). For a given graph \( g \in \{d,\ tt\} \), the node-level hidden representations at time \( t \) are computed as:

\begin{equation}
\mathbf{Z}_t^g = \text{ReLU}\left( \tilde{\mathbf{A}}_t^g \mathbf{H}_t \mathbf{W}^g \right), \quad \forall g \in \{d, tt\}
\end{equation}

Here, \( \tilde{\mathbf{A}}_t^g \in \mathbb{R}^{N_t \times N_t} \) is the normalized adjacency matrix corresponding to graph \( g \), and \( \mathbf{W}^g \in \mathbb{R}^{(F_t + F_s) \times H} \) are graph-specific trainable weight matrices. \( H \) is the hidden dimension of the GCN layer.

To combine the information from the two graph-specific representations, we adopt an attention-based fusion mechanism. The graph-specific outputs are first stacked along a new dimension:

\begin{equation}
\mathbf{Z}_t = \left[ \mathbf{Z}_t^d,\ \mathbf{Z}_t^{tt} \right] \in \mathbb{R}^{N_t \times 2 \times H}
\end{equation}

Let \( \mathbf{w}^g \in \mathbb{R}^{H} \) be the learnable attention weight vector associated with graph \( g \). For each node \( i \in \{1, \dots, N_t\} \), the attention score corresponding to graph \( g \in \{d, tt\} \) is computed as follows:

\begin{equation}
\alpha_{t,i}^g = \frac{\exp\left( \mathbf{Z}_{t,i}^g \cdot \mathbf{w}^g \right)}{\sum_{g' \in \{d, tt\}} \exp\left( \mathbf{Z}_{t,i}^{g'} \cdot \mathbf{w}^{g'} \right)}, \quad \forall g \in \{d, tt\}
\end{equation}

These attention scores represent the relative contribution of each graph in forming the final node representation. The fused node embedding for node \( i \) at time \( t \) is obtained as follows.

\begin{equation}
\mathbf{Z}_{t,i}^{\text{fused}} = \sum_{g \in \{d, tt\}} \alpha_{t,i}^g \mathbf{Z}_{t,i}^g
\end{equation}

By integrating node embeddings from both $G_t^d$ and $G_t^{tt}$ using attention mechanism, the model can dynamically adjust the importance of each graph for accurate prediction.

To capture the temporal dependencies of evacuation traffic, we apply a Long Short-Term Memory (LSTM) network independently to each node across the temporal axis. This fused GCN embedding \(\mathbf{Z}_{t,i}^{\text{fused}} \in \mathbb{R}^{H} \) serves as the input to the downstream LSTM module for temporal modeling. \(\mathbf{Z}_{t,i}^{\text{fused}}\) denote the input to the LSTM for node \( i \) at time \( t \). Let \( \mathbf{h}_{t,i} \in \mathbb{R}^{H} \) represent the corresponding hidden state. The LSTM cell consists of the following components for each node \( i \):

\begin{align}
\mathbf{f}_{t,i} &= \sigma\left( \mathbf{W}_f \mathbf{Z}_{t,i}^{\text{fused}} + \mathbf{U}_f \mathbf{h}_{t-1,i} + \mathbf{b}_f \right) \\
\mathbf{i}_{t,i} &= \sigma\left( \mathbf{W}_i \mathbf{Z}_{t,i}^{\text{fused}} + \mathbf{U}_i \mathbf{h}_{t-1,i} + \mathbf{b}_i \right) \\
\tilde{\mathbf{c}}_{t,i} &= \tanh\left( \mathbf{W}_c \mathbf{Z}_{t,i}^{\text{fused}} + \mathbf{U}_c \mathbf{h}_{t-1,i} + \mathbf{b}_c \right) \\
\mathbf{c}_{t,i} &= \mathbf{f}_{t,i} \odot \mathbf{c}_{t-1,i} + \mathbf{i}_{t,i} \odot \tilde{\mathbf{c}}_{t,i} \\
\mathbf{o}_{t,i} &= \sigma\left( \mathbf{W}_o \mathbf{Z}_{t,i}^{\text{fused}} + \mathbf{U}_o \mathbf{h}_{t-1,i} + \mathbf{b}_o \right) \\
\mathbf{h}_{t,i} &= \mathbf{o}_{t,i} \odot \tanh(\mathbf{c}_{t,i}) 
\end{align}

where, \( \sigma(\cdot) \) denotes the sigmoid activation function, \( \tanh(\cdot) \) is the hyperbolic tangent function, and \( \odot \) denotes element-wise multiplication. \( \mathbf{c}_{t,i} \in \mathbb{R}^{H} \) represent the cell state and \( \mathbf{h}_{t,i} \) is the hidden state. The matrices \( \mathbf{W}_* \in \mathbb{R}^{H \times H} \), \( \mathbf{U}_* \in \mathbb{R}^{H \times H} \), and bias vectors \( \mathbf{b}_* \in \mathbb{R}^{H} \) are learnable parameters shared across all nodes and time steps. The final hidden state \( \mathbf{h}_{t,i} \) of each node is passed through a linear layer to produce the traffic volume predictions for \( p \) future time steps:

\begin{equation}
\hat{\mathbf{y}}_i = \mathbf{W}_{\text{out}} \mathbf{h}_{t,i} + \mathbf{b}_{\text{out}}, \quad \hat{\mathbf{y}}_i \in \mathbb{R}^{p} 
\end{equation}

where, \( \mathbf{W}_{\text{out}} \in \mathbb{R}^{p \times H} \) and \( \mathbf{b}_{\text{out}} \in \mathbb{R}^{p} \) are learnable output projection parameters.

This modeling framework first captures spatial correlations from heterogeneous graph structures and then learn temporal evolution patterns of traffic through the LSTM. The combination improves the accuracy of traffic predictions under rapidly changing and uncertain evacuation conditions.

\subsection{RL-IFSR: Reinforcement Learning-based Intelligent Feature Selection and Ranking}
To assess the contribution of individual input features toward evacuation traffic prediction, we propose an Reinforcement Learning-based Intelligent Feature Selection and Ranking (RL-IFSR). This mechanism leverages a reinforcement learning agent to identify and mask low-utility features during model training, thereby implicitly ranking features by their predictive importance.

We model the feature selection process as a Markov Decision Process (MDP), where a reinforcement learning agent selects features to mask based on the current state of the input. Specifically, we adopt a Double Deep Q-Network (DDQN) algorithm to train the RL agent. The DDQN agent learns a feature-ranking policy that maximizes prediction performance by exploring which features are essential and which can be suppressed with minimal loss.

At each training step, the agent receives a state vector \( \mathbf{s} \in \mathbb{R}^{F_t + F_s} \), defined as follows:

\begin{equation}
\mathbf{s} = \text{mean}_{b,t}(\mathbf{X}_{\text{temp}}) \, \Vert \, \text{mean}_b(\mathbf{X}_{\text{spatial}})
\end{equation}

where \( \mathbf{X}_{\text{temp}} \in \mathbb{R}^{N_t \times T \times F_t} \) are normalized temporal features and \( \mathbf{X}_{\text{spatial}} \in \mathbb{R}^{N_t \times F_s} \) are normalized spatial features. The operator \( \Vert \) denotes vector concatenation across the feature dimension. The means are computed over the batch dimension \( b \) and time steps \( t \), producing a compact representation of the input at each step.

The agent operates over a discrete action space that corresponds to the indices of all input features. At each iteration, the agent selects one feature index \( a \in \mathcal{A} \) to mask, where the action space  \( \mathcal{A} \) is defined as follows:

\begin{equation}
\mathcal{A} = \{0, 1, \dots, F_t + F_s - 1\}
\end{equation}

Here, \( F_t \) and \( F_s \) represent the number of temporal and spatial features, respectively. The first \( F_t \) actions correspond to temporal features, and the remaining \( F_s \) actions correspond to spatial features. If \( a < F_t \), the selected temporal feature is masked; otherwise, the spatial feature with index \( a - F_t \) is masked. This setup allows the agent to isolate and evaluate the influence of each feature in a controlled manner. Only one feature is masked at a time, enabling an estimation of its marginal contribution to the model’s predictive accuracy. 

The RL agent apply binary masks to the temporal and spatial input tensors. Let \( \mathbf{m}_{\text{temp}} \in \mathbb{R}^{F_t} \) and \( \mathbf{m}_{\text{spatial}} \in \mathbb{R}^{F_s} \) denote binary masks initialized with all ones. The selected feature index \( a \in \mathcal{A} \) determines which feature to suppress by setting the corresponding mask value to zero:
\begin{align}
&\mathbf{m}_{\text{temp}}[a] = 0, \quad \text{if } a < F_t \\
&\mathbf{m}_{\text{spatial}}[a - F_t] = 0, \quad \text{if } a \geq F_t
\end{align}

These masks are then applied to the input tensors via element-wise multiplication, effectively zeroing out the selected feature:

\begin{align}
&\tilde{\mathbf{X}}_{\text{temp}} = \mathbf{X}_{\text{temp}} \odot \mathbf{m}_{\text{temp}} \\
&\tilde{\mathbf{X}}_{\text{spatial}} = \mathbf{X}_{\text{spatial}} \odot \mathbf{m}_{\text{spatial}}
\end{align}

Here, \( \tilde{\mathbf{X}}_{\text{temp}} \) and \( \tilde{\mathbf{X}}_{\text{spatial}} \) denote the masked versions of the temporal and spatial features, respectively. They are the input features (see Equation~\ref{eq:input_features} for the dynamic multi-graph fusion model described in Section~\ref{sec:dmf}.

We train the RL agent by adopting DDQN algorithm. In our formulation, let \( Q(\mathbf{s}, a; \theta) \) denote the Q-value predicted by the online network with parameters \( \theta \), for a given state \( \mathbf{s} \in \mathbb{R}^{F_t + F_s} \) and action \( a \in \mathcal{A} \). At each training iteration, the agent transitions to a new state \( \mathbf{s}' \), receives a reward \( r \), and updates the Q-value estimate using the DDQN target network, defined as follows:

\begin{equation} \label{eq:ddqn}
y = r + \gamma \cdot Q\left(\mathbf{s}', \underset{a'}{\arg\max} \, Q(\mathbf{s}', a'; \theta); \theta^{-}\right)
\end{equation}
Here, the online network \( Q(\cdot; \theta) \) is used to select the next action \( a' \) that maximizes the predicted Q-value in the next state \( \mathbf{s}' \), while the target network \( Q(\cdot; \theta^{-}) \) is used to evaluate the Q-value of that selected action. The discount factor \( \gamma \in [0, 1] \) controls the contribution of future rewards.

The Q-learning loss is calculated as the squared difference between the target \( y \) and the current estimate $Q(s,a;\theta$). It is defined as follows,
\begin{equation}
\mathcal{L}_{\text{DDQN}} = \left( y - Q(\mathbf{s}, a; \theta) \right)^2
\end{equation}

To improve the stability and efficiency of the learning process, we employ a prioritized experience replay buffer \cite{schaul2016prioritized} that stores past transitions in the form of tuples \( (\mathbf{s}, a, r, \mathbf{s}') \). Instead of sampling transitions uniformly, the agent prioritizes experiences that are deemed more informative for learning, allowing the model to focus on transitions that have higher potential to improve prediction accuracy and policy updates. This prioritization enables the agent to focus on transitions that are likely to reduce prediction errors faster, improving convergence speed and sample efficiency. The sampled experiences are then used to compute the Double DQN target and perform Q-network updates as described in Equation~\ref{eq:ddqn}.

To balance the trade-off between exploration (trying new actions) and exploitation (using the best-known action with highest Q-value), the agent adopts an $\epsilon$-greedy strategy during training. This decision-making process is formalized as follows,
\begin{equation}
a =
\begin{cases}
\text{random action}, & \text{with probability } \epsilon \\
\arg\max_{a} Q(\mathbf{s}, a; \theta), & \text{with probability } 1 - \epsilon
\end{cases}
\end{equation}
where, $\epsilon$ represents the exploration probability i.e., how often the DDQN agent chooses a random action instead of the best-known action with the highest Q-value. At the beginning of training, $\epsilon$ is high (for our case, $\epsilon_{min} = 1.0$), so the agent explores by randomly masking different features to gather diverse experience. Over time, $\epsilon$ decays (decay factor = 0.995), so the agent starts to rely more on what it has learned. By the end of training, $\epsilon$ becomes very low (for our case, $\epsilon_{min} = 0.05$, so the agent mostly exploits by consistently masking least important features for better prediction.

By tracking the frequency and impact of selected actions during training, we interpret the policy learned by DDQN agent as an implicit mechanism for feature ranking. The agent learns to selectively mask features whose removal has the least detrimental effect on prediction performance. Over time, it develops a preference for retaining more informative features, while consistently suppressing those deemed less useful. As a result, features that are rarely masked i.e., whose masking consistently leads to higher prediction error — are considered to be more important. Conversely, features that are frequently masked are likely to carry less predictive value. 

\begin{figure*}[htbp]
    \centering
    \includegraphics[width=0.98\linewidth]{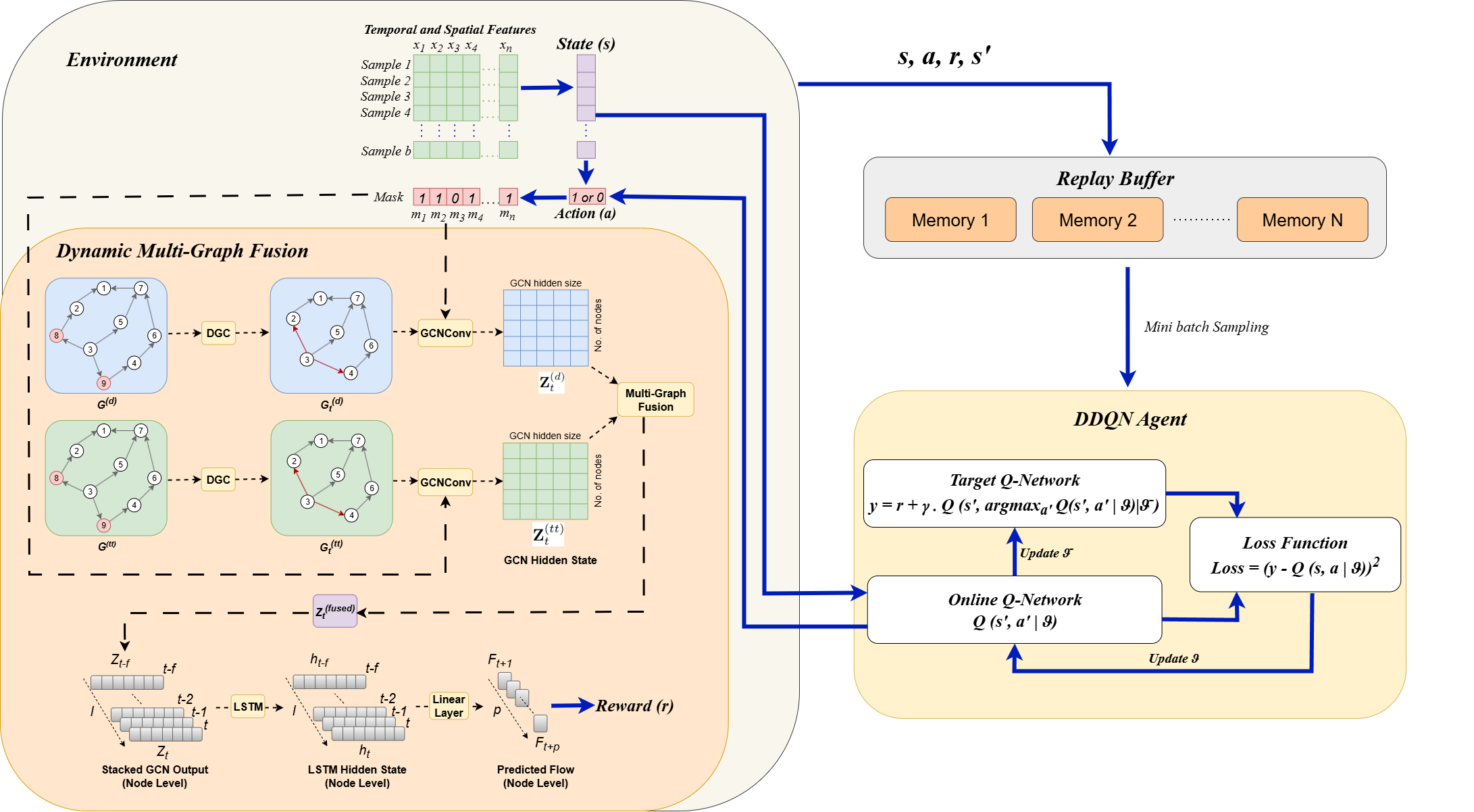}
    \caption{The Proposed RL-Guided Dynamic Multi-Graph Fusion (RL-DMF) Framework}
    \label{fig:rl-dmf}
\end{figure*}

The RL agent receives a scalar reward \( r \in \mathbb{R} \) based on the negative prediction loss of the underlying dynamic multi-graph fusion model (described in previous section). This reward serves as feedback indicating the effectiveness of the selected feature masking action. Specifically, the reward is computed as:
\begin{equation}
r = -\mathcal{L}(\hat{\mathbf{y}}, \mathbf{y})
\end{equation}
where \( \hat{\mathbf{y}} \in \mathbb{R}^{N_t \times p} \) denotes the predicted traffic volume for all nodes over the output sequence length \( p \), and \( \mathbf{y} \) is the actual traffic volume. The loss function \( \mathcal{L} \) for our case, is the mean squared error (MSE). A higher prediction accuracy (i.e., lower loss) results in a more positive reward, guiding the agent to favor feature selections that support better model performance. Conversely, if masking a particular feature degrades the prediction, the corresponding reward becomes more negative. This formulation directly couples the agent’s behavior with the predictive success of the traffic forecasting model, enabling the agent to learn a feature masking policy that enhances the performance over time.

Figure~\ref{fig:rl-dmf} shows the architecture of the proposed RL-DMF framework which integrates Reinforcement Learning-based Intelligent Feature Selection and Ranking (RL-IFSR) with a Dynamic Multi-Graph Fusion (DMF) model. The fusion of graphs allows the model to learn richer representations of the dynamic traffic network, while the RL-guided feature selection enhances the interpretability and generalizability of the model.

\subsection{Model Training, Baselines, and Evaluation Metrics} \label{sec:MBE}
To evaluate the effectiveness of the proposed framework, we trained the model using historical traffic and evacuation data collected from multiple past hurricanes. The dataset comprises hourly records of traffic flow, speed, incident and evacuation attributes, etc. We used 90\% of the data for training and remaining 10\% for validation. We used the ADAM optimizer \cite{kingma2015adam} with stochastic gradient descent (SGD) to update model parameters. Training was conducted on the Newton High-Performance Computing (HPC) cluster at the University of Central Florida \cite{ucf_newton}, utilizing GPU resources of 164~GB shared GPU memory. The training process took approximately 2.5 hours to complete. Then, we tested the model on Hurricane Milton's two days evacuation period: October 7 and 8, 2024. We used 6-hour historical input window (\( l = 6 \)) to predict traffic volume for the subsequent 6-hour horizon (\( p = 6 \)).

To evaluate the performance of RL-DMF framework, we compare it with several baseline models for evacuation traffic prediction. These baseline models are as follows.

\begin{itemize}
    \item \textbf{LSTM}: A standard Long Short-Term Memory network that models temporal dependencies in the traffic data without considering spatial relationships.
    
    \item \textbf{CNN-LSTM}: A hybrid model that first applies convolutional layers to capture short-range spatial patterns, followed by LSTM layers to learn temporal dependencies.
    
    \item \textbf{Static GCN-LSTM}: A graph-based baseline where a fixed, static road network graph is used to perform spatial learning via Graph Convolutional Networks (GCNs), followed by LSTM layers for temporal modeling. The graph structure does not change over time.
    
    \item \textbf{Dynamic GCN-LSTM (Distance-based)}: A dynamic graph learning model in which the graph structure evolves over time, and the edge weights are computed based on normalized physical distances between detectors. This model captures temporal variations in node availability and spatial connectivity.
    
    \item \textbf{Dynamic GCN-LSTM (Travel-time-based)}: Similar to the previous model, this baseline uses time-varying dynamic graphs; the edge weights are derived from normalized travel times between detectors, thereby incorporating traffic dynamics into the graph structure.
\end{itemize}

All models were implemented using the PyTorch Geometric library in Python \cite{fey2019graph}. To assess model performance, we employed four evaluation metrics: Root Mean Square Error (RMSE), Mean Absolute Error (MAE), Mean Absolute Percentage Error (MAPE), and the coefficient of determination (R\textsuperscript{2}). These metrics are defined as follows.

\begin{equation}
\text{RMSE} = 
\sqrt{\frac{1}{N} \sum_{i=1}^{N} \left(F_{\text{actual},i} - F_{\text{predicted},i}\right)^2}
\label{eq:rmse}
\end{equation}

\begin{equation}
\text{MAE} = \frac{1}{N} \sum_{i=1}^{N} \left|F_{\text{actual},i} - F_{\text{predicted},i}\right|
\label{eq:mae}
\end{equation}

\begin{equation}
\text{MAPE} = \frac{1}{N} \sum_{i=1}^{N} \left| \frac{F_{\text{actual},i} - F_{\text{predicted},i}}{F_{\text{actual},i}} \right| \times 100\%
\label{eq:mape}
\end{equation}

\begin{equation}
R^2 = 1 - \frac{\sum_{i=1}^{N} \left(F_{\text{actual},i} - F_{\text{predicted},i}\right)^2}
{\sum_{i=1}^{N} \left(F_{\text{actual},i} - \overline{F}_{\text{actual}}\right)^2}
\label{eq:r2}
\end{equation}

where:
 \( N \) is the total number of samples;
 \( F_{\text{actual},i} \) represents actual flow for \( i \)-th sample;
 \( F_{\text{predicted},i} \) represents predicted flow for \( i \)-th sample; and
 \( \overline{F}_{\text{actual}} \) is the mean of the actual flows.

\section{Results} \label{sec:results}
\begin{figure}[htbp]
    \centering
    \includegraphics[width=0.95\linewidth]{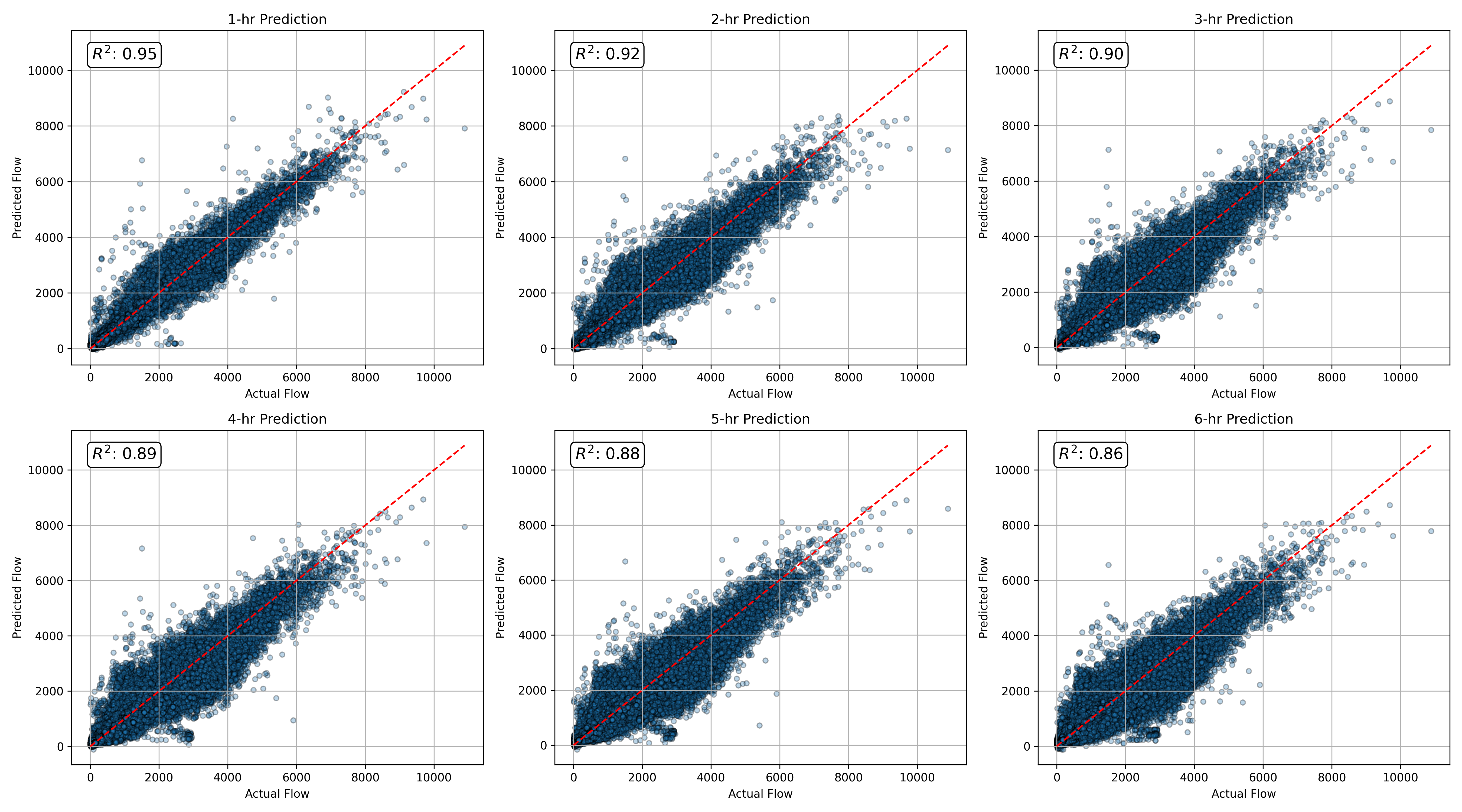}
    \caption{Comparison between actual and predicted traffic flows for 1-hour to 6-hour prediction horizons during Milton}
    \label{fig:flow_scatter}
\end{figure}

We predicted the traffic flows during the evacuation period of Milton. Table~\ref{tab:evacuation_metrics_milton} presents the performance of the RL-DMF model for Hurricane Milton across prediction horizons ranging from 1 to 6 hours. The model achieves a 95\% accuracy with RMSE of 293.9 when predicting the next 1-hour traffic flow. The results show that error metrics tend to increase with longer prediction horizons, which is expected in multi-step forecasting tasks. Further, Figure~\ref{fig:flow_scatter} illustrates the scatter plots of actual and predicted traffic flows across prediction horizons from 1 to 6 hours. The plots also show a similar pattern; as the prediction horizon increases, the spread of the points becomes wider, indicating reduced prediction accuracy. Despite this, the R\textsuperscript{2} values remain relatively high (above 0.86 for all horizons), confirming the model's strong predictive power even at longer prediction horizons and indicating it's ability to capture long-term dependency. Overall, on average, the model shows robust performance with an accuracy of 90\% and an RMSE of 426.4 considering results of all prediction horizons (from 1-hour to 6-hour prediction horizons).

\begin{table}[htbp]
\centering
\caption{Evacuation Traffic Prediction Performance for Milton}
\label{tab:evacuation_metrics_milton}
{\scriptsize (\textit{Minimum flow = 6, Maximum flow = 10889, Mean flow = 1838, Median flow = 1503})}
\begin{tabularx}{\linewidth}{l XXXX}
\toprule
\textbf{Prediction Horizon} & \textbf{RMSE} & \textbf{MAE} & \textbf{MAPE} & \textbf{R\textsuperscript{2}} \\
\midrule
1-hour & 293.9 & 189.5 & 17.9 & 0.95 \\
2-hour & 380.0 & 248.4 & 20.8 & 0.92 \\
3-hour & 430.6 & 285.1 & 23.9 & 0.90 \\
4-hour & 455.0 & 304.9 & 27.2 & 0.89 \\
5-hour & 471.4 & 320.3 & 29.7 & 0.88 \\
6-hour & 495.3 & 338.2 & 31.7 & 0.86 \\
\midrule
\textbf{Overall} & \textbf{426.4} & \textbf{281.1} & \textbf{25.2} & \textbf{0.90} \\
\bottomrule
\end{tabularx}
\end{table}

Table~\ref{tab:baseline_comparison} shows the comparison of performance of RL-DMF model with several baseline models. Traditional LSTM and CNN-LSTM architectures show inferior performance with higher error values compared to other models. The Static GCN-LSTM and Dynamic GCN-LSTM models (using either distance-based or travel-time-based graph) perform better than previous LSTM-based models, benefiting from their ability to capture spatial dependencies. However, the RL-DMF model outperforms all baseline models for all evaluation metrics. These results demonstrate the effectiveness of integrating dynamic graph fusion with reinforcement learning-based feature selection for a better prediction.

\begin{table}[htbp]
\centering
\caption{Performance Comparison of RL-DMF with Baseline Models for 1-hour to 6-hour prediction horizons (Milton)}
\label{tab:baseline_comparison}
{\scriptsize (\textit{Minimum flow = 6, Maximum flow = 10889, Mean flow = 1838, Median flow = 1503})}
{\small\begin{tabularx}{\linewidth}{l XXXXX}
\toprule
\textbf{Model} & \textbf{Horizon} & \textbf{RMSE} & \textbf{MAE} & \textbf{MAPE} & \textbf{R\textsuperscript{2}} \\
\midrule

\multirow{7}{*}{LSTM}
  & 1-hour & 308.1 & 206.3 & 21.8 & 0.95 \\
  & 2-hour & 411.9 & 276.3 & 24.3 & 0.91 \\
  & 3-hour & 476.1 & 326.1 & 29.0 & 0.88 \\
  & 4-hour & 517.5 & 359.1 & 32.4 & 0.85 \\
  & 5-hour & 549.6 & 380.3 & 34.2 & 0.83 \\
  & 6-hour & 571.7 & 397.3 & 36.2 & 0.82 \\
  & \textbf{Overall} & \textbf{481.0} & \textbf{324.2} & \textbf{29.6} & \textbf{0.87} \\
\midrule

\multirow{7}{*}{CNN-LSTM}
  & 1-hour & 389.0 & 265.2 & 21.3 & 0.92 \\
  & 2-hour & 473.3 & 317.5 & 24.8 & 0.88 \\
  & 3-hour & 538.9 & 358.9 & 27.2 & 0.84 \\
  & 4-hour & 576.1 & 385.1 & 30.1 & 0.82 \\
  & 5-hour & 594.0 & 401.3 & 32.8 & 0.81 \\
  & 6-hour & 616.9 & 417.2 & 34.9 & 0.79 \\
  & \textbf{Overall} & \textbf{537.1} & \textbf{357.5} & \textbf{28.5} & \textbf{0.84} \\
\midrule

\multirow{7}{*}{Static GCN-LSTM}
  & 1-hour & 296.8 & 189.4 & 16.4 & 0.95 \\
  & 2-hour & 411.2 & 261.1 & 20.5 & 0.91 \\
  & 3-hour & 478.1 & 309.2 & 24.7 & 0.88 \\
  & 4-hour & 513.4 & 337.2 & 26.7 & 0.86 \\
  & 5-hour & 541.9 & 361.7 & 29.8 & 0.84 \\
  & 6-hour & 583.2 & 393.5 & 35.9 & 0.81 \\
  & \textbf{Overall} & \textbf{480.1} & \textbf{308.7} & \textbf{25.6} & \textbf{0.87} \\
\midrule

\multirow{7}{*}{\parbox{3cm}{\centering Dynamic\\GCN-LSTM\\(Distance-based)}}
  & 1-hour & 304.6 & 195.8 & 16.8 & 0.95 \\
  & 2-hour & 416.8 & 270.2 & 21.8 & 0.91 \\
  & 3-hour & 480.9 & 311.9 & 24.8 & 0.88 \\
  & 4-hour & 518.1 & 343.3 & 27.6 & 0.85 \\
  & 5-hour & 551.6 & 372.9 & 31.2 & 0.83 \\
  & 6-hour & 572.0 & 389.4 & 33.7 & 0.82 \\
  & \textbf{Overall} & \textbf{482.6} & \textbf{313.9} & \textbf{26.0} & \textbf{0.87} \\
\midrule

\multirow{7}{*}{\parbox{3cm}{\centering Dynamic\\GCN-LSTM\\(Travel-Time-based)}}
  & 1-hour & 293.4 & 192.1 & 18.2 & 0.95 \\
  & 2-hour & 400.2 & 258.9 & 21.1 & 0.92 \\
  & 3-hour & 470.3 & 309.9 & 24.7 & 0.88 \\
  & 4-hour & 515.0 & 346.6 & 28.7 & 0.86 \\
  & 5-hour & 543.5 & 367.6 & 31.7 & 0.84 \\
  & 6-hour & 570.1 & 386.2 & 33.1 & 0.82 \\
  & \textbf{Overall} & \textbf{474.9} & \textbf{310.2} & \textbf{26.2} & \textbf{0.88} \\
\midrule

\multirow{7}{*}{\parbox{3.1cm}{\centering \textbf{RL-DMF}\\\textbf{(Proposed Model)}}}
  & 1-hour & 293.9 & 189.5 & 17.9 & 0.95 \\
  & 2-hour & 380.0 & 248.4 & 20.8 & 0.92 \\
  & 3-hour & 430.6 & 285.1 & 23.9 & 0.90 \\
  & 4-hour & 455.0 & 304.9 & 27.2 & 0.89 \\
  & 5-hour & 471.4 & 320.3 & 29.7 & 0.88 \\
  & 6-hour & 495.3 & 338.2 & 31.7 & 0.86 \\
  & \textbf{Overall} & \textbf{426.4} & \textbf{281.1} & \textbf{25.2} & \textbf{0.90} \\
\bottomrule
\end{tabularx}}
\end{table}

\begin{figure}[htbp]
    \centering
    \includegraphics[width=0.9\linewidth]{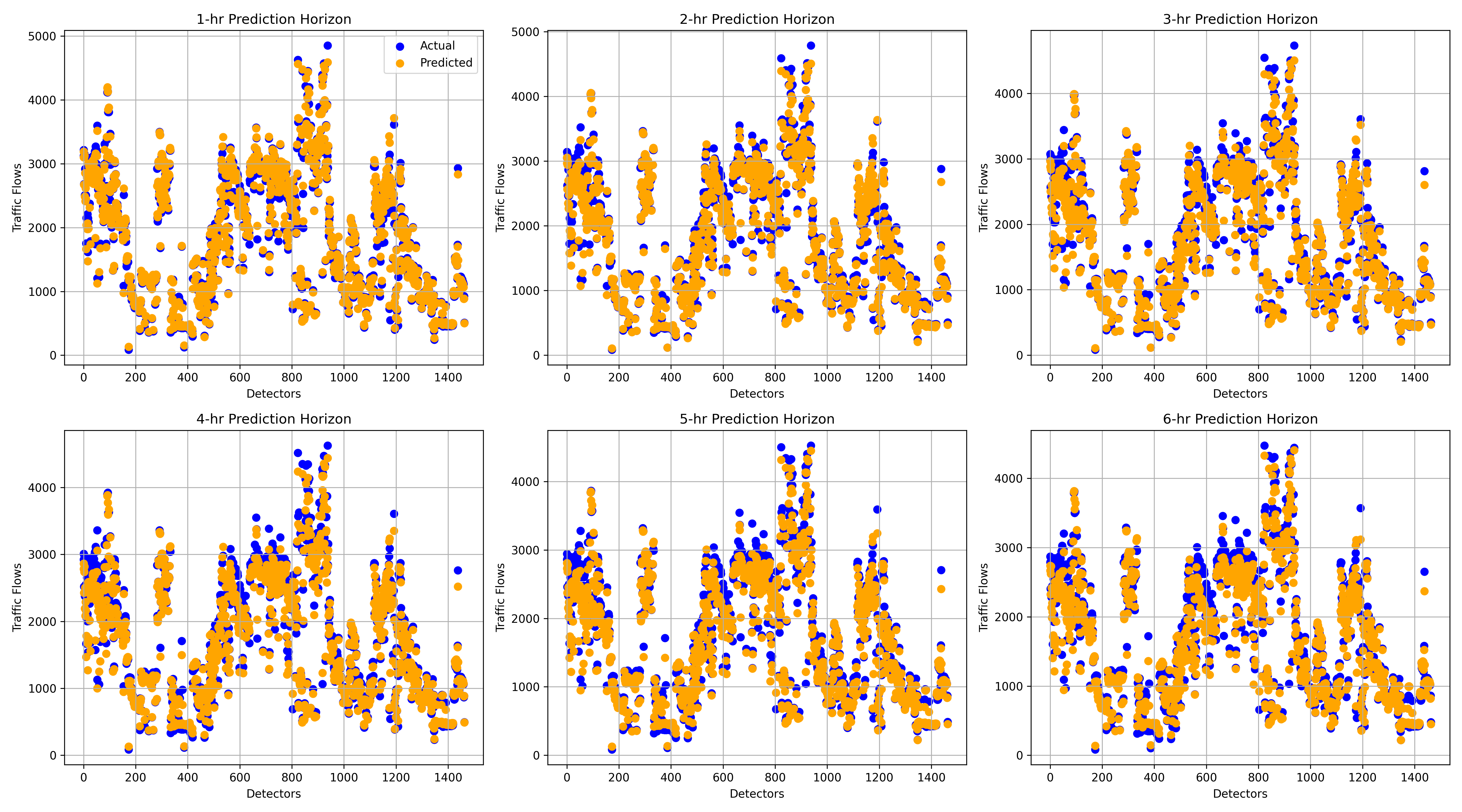}
    \caption{Detector-wise comparison of average actual and predicted flow for 1-hour to 6-hour prediction horizons during Milton}
    \label{fig:flow_dist}
\end{figure}

We also visualize the detector-wise average of actual and predicted traffic flows over  2-day evacuation period during Milton (see Figure~\ref{fig:flow_dist}). This plot provides insights into the model’s generalization capability at a detector level. Each detector may reflect distinct traffic dynamics, and comparing average flows allows us to evaluate how well the model adapts to these localized patterns in a dynamic network. As evident in Figure~\ref{fig:flow_dist}, while a few detectors show noticeable discrepancies (e.g., blue dots deviating from the orange dots), the majority exhibit strong alignment between actual and predicted flows. This result highlights the model’s generalizability across individual detectors despite the dynamic nature of the network, where detector availability varies over time due to outages, maintenance, or disruptions.

To evaluate the generalizability of RL-DMF model across different unseen hurricane, we trained a separate model and tested it on Hurricane Ian—a rapidly intensifying storm that made landfall in Florida in 2022. The model's performance on Ian is also provided in Table~\ref{tab:evacuation_metrics_ian}, showing an overall predictive accuracy of 90\% with an RMSE of 409.7 across the 1-hour to 6-hour prediction horizons. When compared to the results for Hurricane Milton, Ian's model yields lower RMSE values. A likely explanation is that the scale of evacuation during Milton was significantly larger, with approximately 5.5 million people under evacuation orders, compared to 2.5 million during Ian \cite{nyt2024hurricane, foxweather2022ian}. As a result, higher traffic volumes were observed during Milton (mean flow = 1838) than Ian (mean flow = 1561), leading to relatively lower prediction errors for Ian. Nevertheless, the model's overall MAPE for Ian (33.0\%) is higher than that for Milton (25.2\%). Overall, the model demonstrates strong generalizability across hurricanes, maintaining high performance not only for Milton but also for another unseen event like Hurricane Ian.

\begin{table}[htbp]
\centering
\caption{Evacuation Traffic Prediction Performance for Hurricane Ian}
\label{tab:evacuation_metrics_ian}
{\scriptsize (\textit{Minimum flow = 1, Maximum flow = 10240, Mean flow = 1561, Median flow = 1161})}
\begin{tabularx}{\linewidth}{l XXXX}
\toprule
\textbf{Prediction Horizon} & \textbf{RMSE} & \textbf{MAE} & \textbf{MAPE} & \textbf{R\textsuperscript{2}} \\
\midrule
1-hour & 270.0 & 170.5 & 21.1 & 0.96 \\
2-hour & 356.7 & 228.5 & 28.6 & 0.92 \\
3-hour & 415.1 & 266.2 & 32.7 & 0.90 \\
4-hour & 451.2 & 289.3 & 36.2 & 0.87 \\
5-hour & 462.1 & 299.6 & 37.0 & 0.87 \\
6-hour & 466.7 & 312.1 & 42.3 & 0.86 \\
\midrule
\textbf{Overall} & \textbf{409.7} & \textbf{261.1} & \textbf{33.0} & \textbf{0.90} \\
\bottomrule
\end{tabularx}
\end{table}

Given that both Hurricane Milton and Hurricane Ian made landfall along Florida’s west coast, the primary evacuation corridors were I-4 eastbound and I-75 northbound \cite{rafi2025dynamic}. To assess model performance under extreme evacuation conditions, we evaluated the model on these two critical highways. As shown in Table~\ref{tab:road_specific_metrics}, the model’s predictive accuracy on I-4 and I-75 is slightly lower than the network-wide performance for both hurricanes. However, the model maintains robust and consistent accuracy on these routes, which is evident by relatively stable MAPE values. Notably, the model achieved a MAPE as low as 11.4\% for 1-hour traffic prediction on I-4 during Hurricane Milton, and 12.1\% for 1-hour traffic prediction on I-4 during Hurricane Ian. Overall, the model demonstrates better MAPE scores for I-4 and I-75 during Milton than during Ian. This highlights the model’s resilience under more severe evacuation conditions associated with Milton. Despite the higher traffic volume and intensity, the proposed model performs reliably better on these major evacuation routes.

\begin{table}[htbp]
\centering
\caption{Highway specific Prediction Performance for Hurricane Milton and Ian}
\label{tab:road_specific_metrics}
\begin{tabularx}{\linewidth}{l|XXXX|XXXX}
\toprule
\multirow{2}{*}{\textbf{Horizon}} & \multicolumn{4}{c|}{\textbf{I-4 (Milton)}} & \multicolumn{4}{c}{\textbf{I-4 (Ian)}} \\
\cmidrule{2-9}
 & \textbf{RMSE} & \textbf{MAE} & \textbf{MAPE} & \textbf{R\textsuperscript{2}} & \textbf{RMSE} & \textbf{MAE} & \textbf{MAPE} & \textbf{R\textsuperscript{2}} \\
\midrule
1-hour & 322.1 & 230.4 & 11.4 & 0.94 & 299.8 & 203.3 & 12.1 & 0.95 \\
2-hour & 453.2 & 313.3 & 14.8 & 0.89 & 410.6 & 274.7 & 16.4 & 0.91 \\
3-hour & 533.4 & 377.3 & 18.2 & 0.84 & 485.1 & 325.7 & 19.2 & 0.88 \\
4-hour & 560.9 & 403.0 & 20.0 & 0.83 & 524.6 & 354.8 & 20.6 & 0.86 \\
5-hour & 593.3 & 438.7 & 23.1 & 0.81 & 548.4 & 373.1 & 21.6 & 0.85 \\
6-hour & 620.1 & 475.2 & 28.1 & 0.79 & 566.4 & 388.7 & 24.8 & 0.83 \\
\midrule
Overall & 523.6 & 373.0 & 19.3 & 0.85 & 481.4 & 320.0 & 19.1 & 0.88 \\
\midrule
\multirow{2}{*}{\textbf{Horizon}} & \multicolumn{4}{c|}{\textbf{I-75 (Milton)}} & \multicolumn{4}{c}{\textbf{I-75 (Ian)}} \\
\cmidrule{2-9}
& \textbf{RMSE} & \textbf{MAE} & \textbf{MAPE} & \textbf{R\textsuperscript{2}} & \textbf{RMSE} & \textbf{MAE} & \textbf{MAPE} & \textbf{R\textsuperscript{2}} \\
\midrule
1-hour & 332.4 & 222.2 & 21.8 & 0.94 & 292.1 & 195.2 & 18.5 & 0.95 \\
2-hour & 424.2 & 287.6 & 23.1 & 0.90 & 398.0 & 272.1 & 24.9 & 0.90 \\
3-hour & 483.8 & 331.7 & 25.7 & 0.87 & 465.5 & 321.8 & 30.0 & 0.87 \\
4-hour & 517.3 & 359.0 & 29.2 & 0.85 & 507.5 & 352.6 & 34.3 & 0.84 \\
5-hour & 541.3 & 379.9 & 33.3 & 0.83 & 524.8 & 370.0 & 38.0 & 0.83 \\
6-hour & 577.5 & 405.9 & 34.7 & 0.81 & 543.0 & 394.6 & 46.5 & 0.81 \\
\midrule
Overall & 486.2 & 331.0 & 27.9 & 0.87 & 463.4 & 317.7 & 32.0 & 0.87 \\
\bottomrule
\end{tabularx}
\end{table}

\begin{figure}[htbp]
    \centering
    \includegraphics[width=0.9\linewidth]{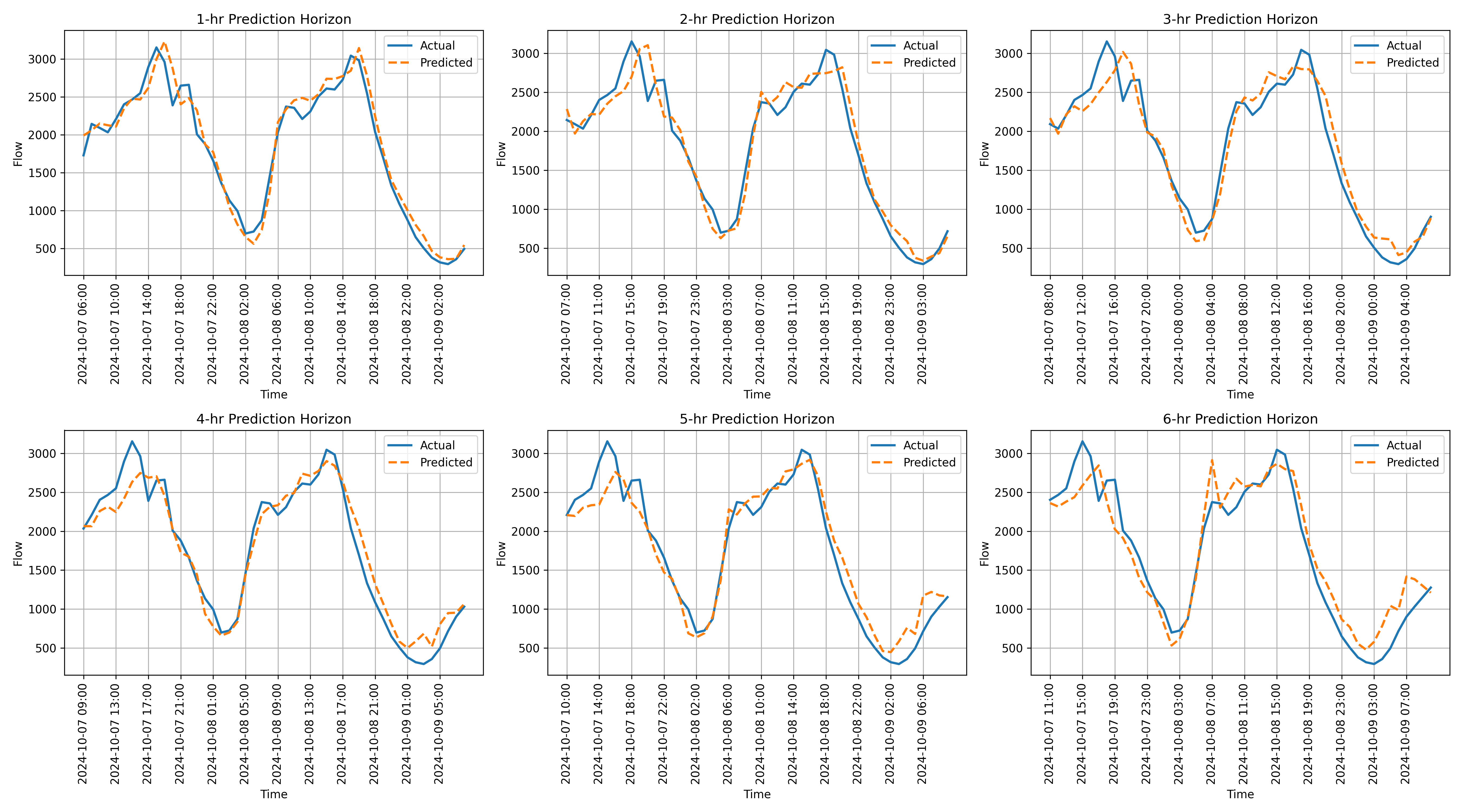}
    \caption{Model performance of an I-4 detector for 6-hour prediction horizon during Milton}
    \label{fig:flow_line_i4}
\end{figure}

\begin{figure}[htbp]
    \centering
    \includegraphics[width=0.9\linewidth]{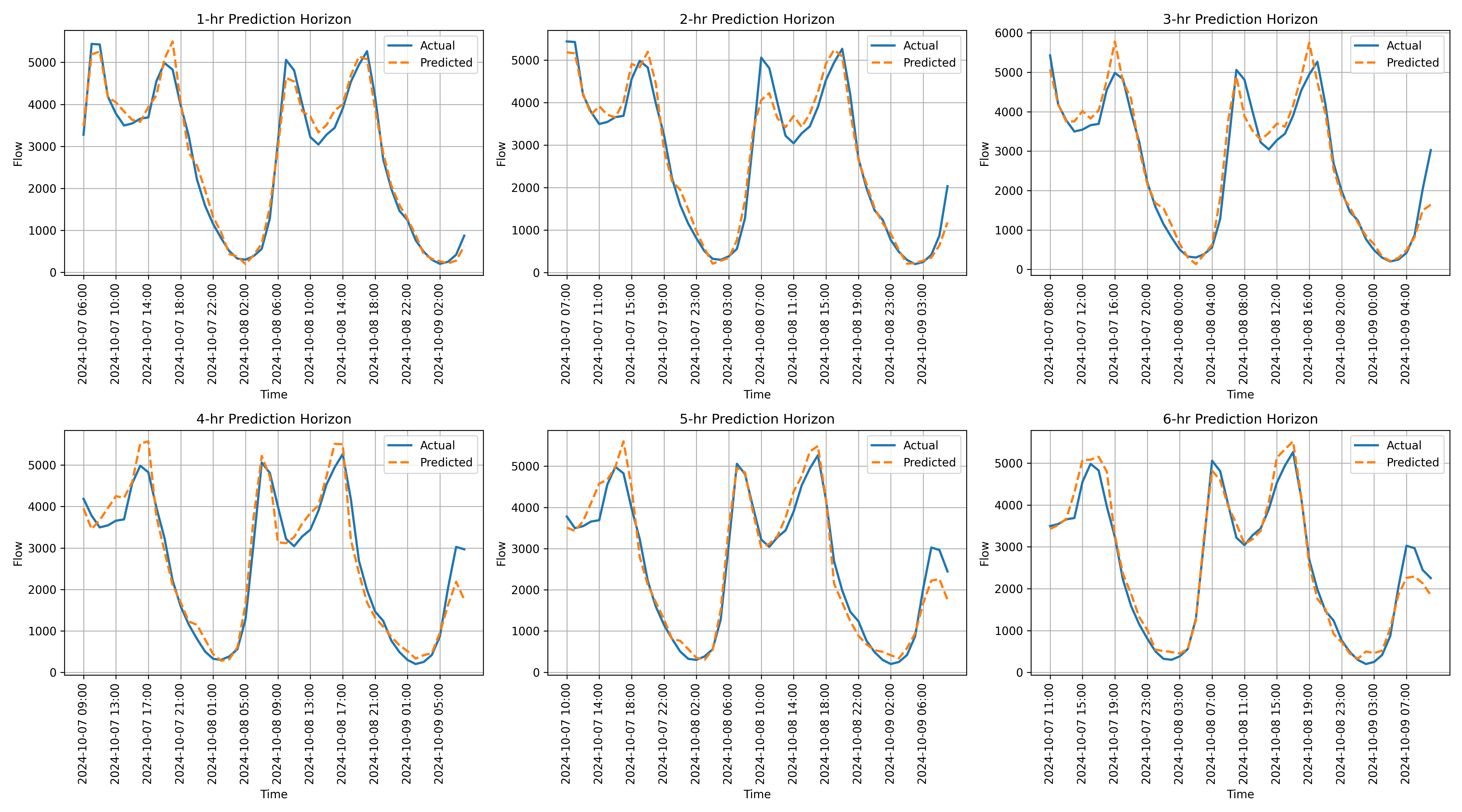}
    \caption{Model performance of an I-75 detector for 6-hour prediction horizon during Milton}
    \label{fig:flow_line_i75}
\end{figure}

Further analysis was conducted by plotting the predicted and actual traffic flows for an individual detector located on I-4 and I-75 during Hurricane Milton under a 6-hour prediction horizon (see Figures~\ref{fig:flow_line_i4} and \ref{fig:flow_line_i75}). For both detectors, the model closely follows the actual traffic patterns, especially for shorter horizons, accurately capturing key flow characteristics such as peaks and troughs. Sometimes, the model deviates slightly from the actual peak traffic volumes on both I-4 and I-75. Nevertheless, the model effectively captures the overall evacuation flow dynamics on both highways.

\begin{figure}[htbp]
    \centering

    \begin{subfigure}[b]{0.3\linewidth}
        \centering
        \includegraphics[width=\linewidth]{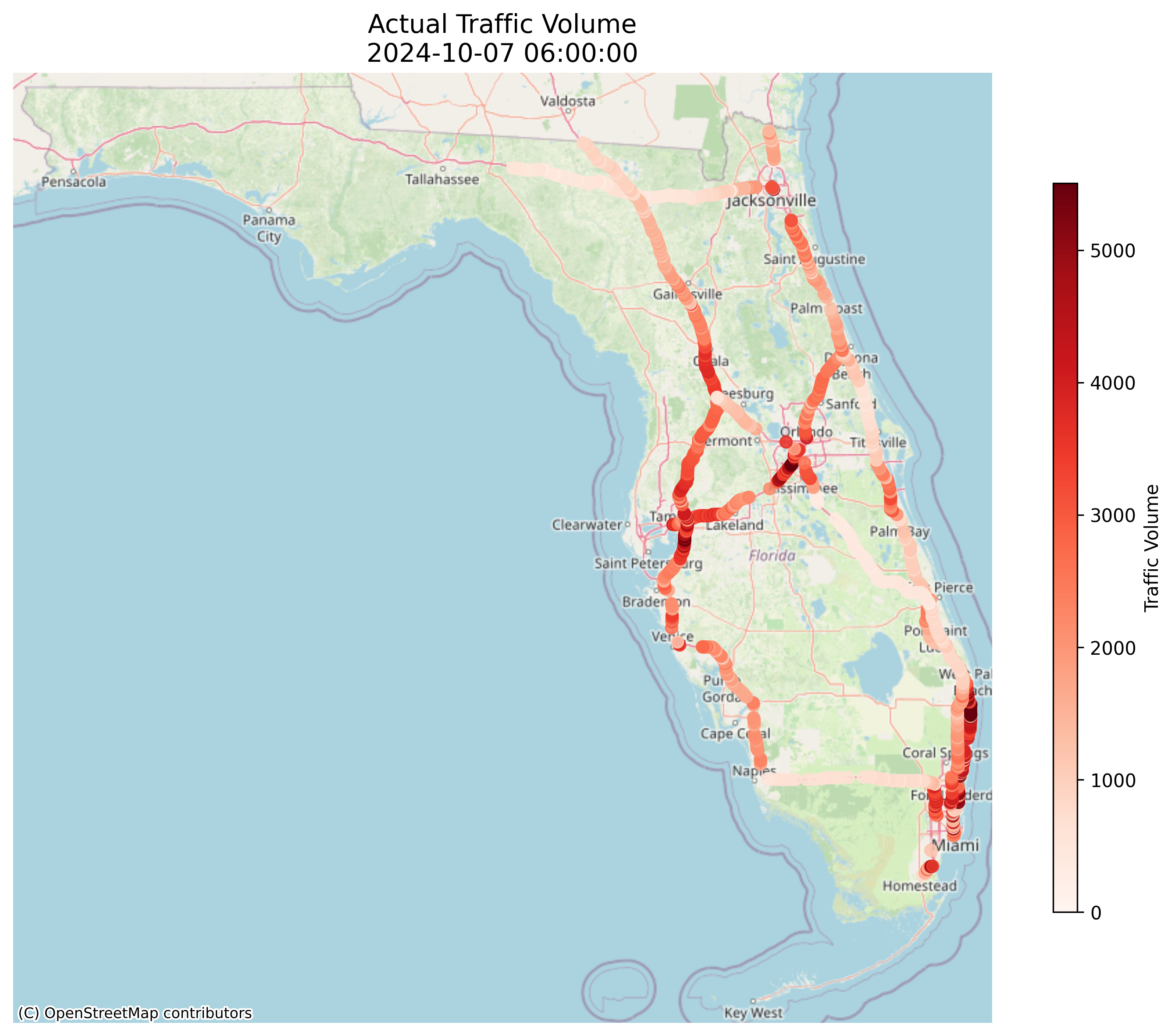}
        \caption{2024-10-07 06:00:00}
    \end{subfigure}
    \hspace{0.02\linewidth}
    \begin{subfigure}[b]{0.3\linewidth}
        \centering
        \includegraphics[width=\linewidth]{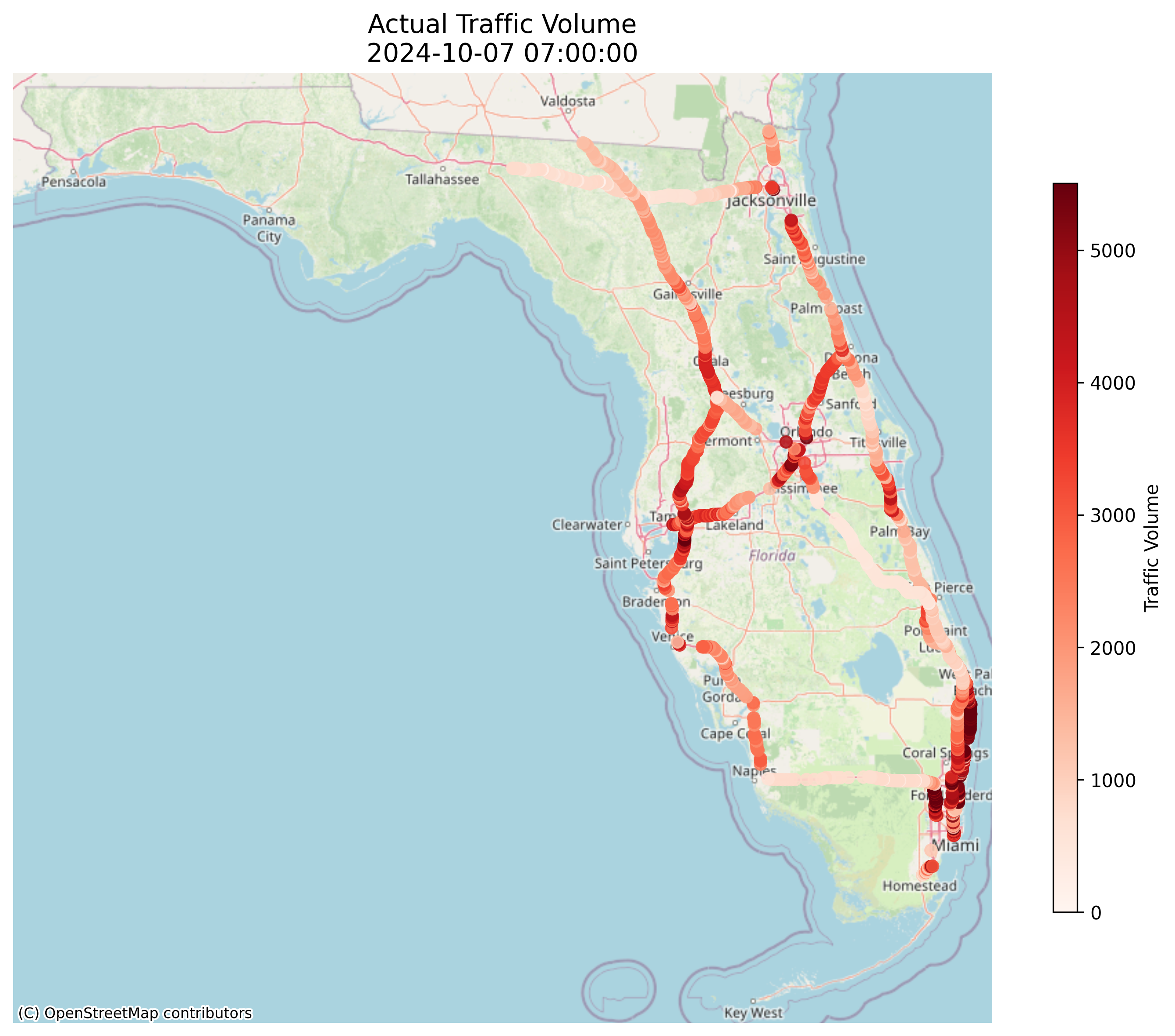}
        \caption{2024-10-07 07:00:00}
    \end{subfigure}
    \hspace{0.02\linewidth}
    \begin{subfigure}[b]{0.3\linewidth}
        \centering
        \includegraphics[width=\linewidth]{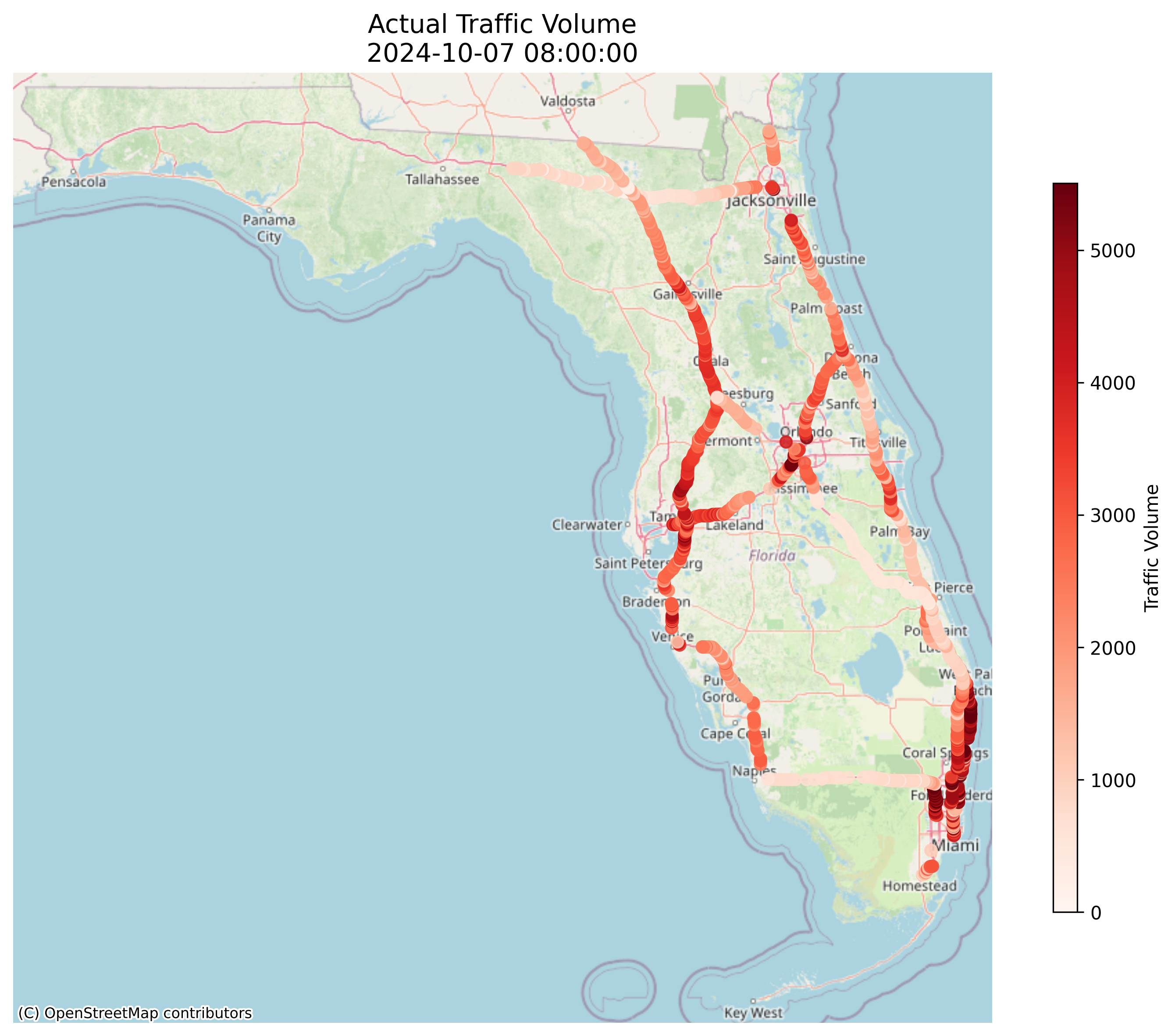}
        \caption{2024-10-07 08:00:00}
    \end{subfigure}

    \vspace{0.3cm}

    \begin{subfigure}[b]{0.3\linewidth}
        \centering
        \includegraphics[width=\linewidth]{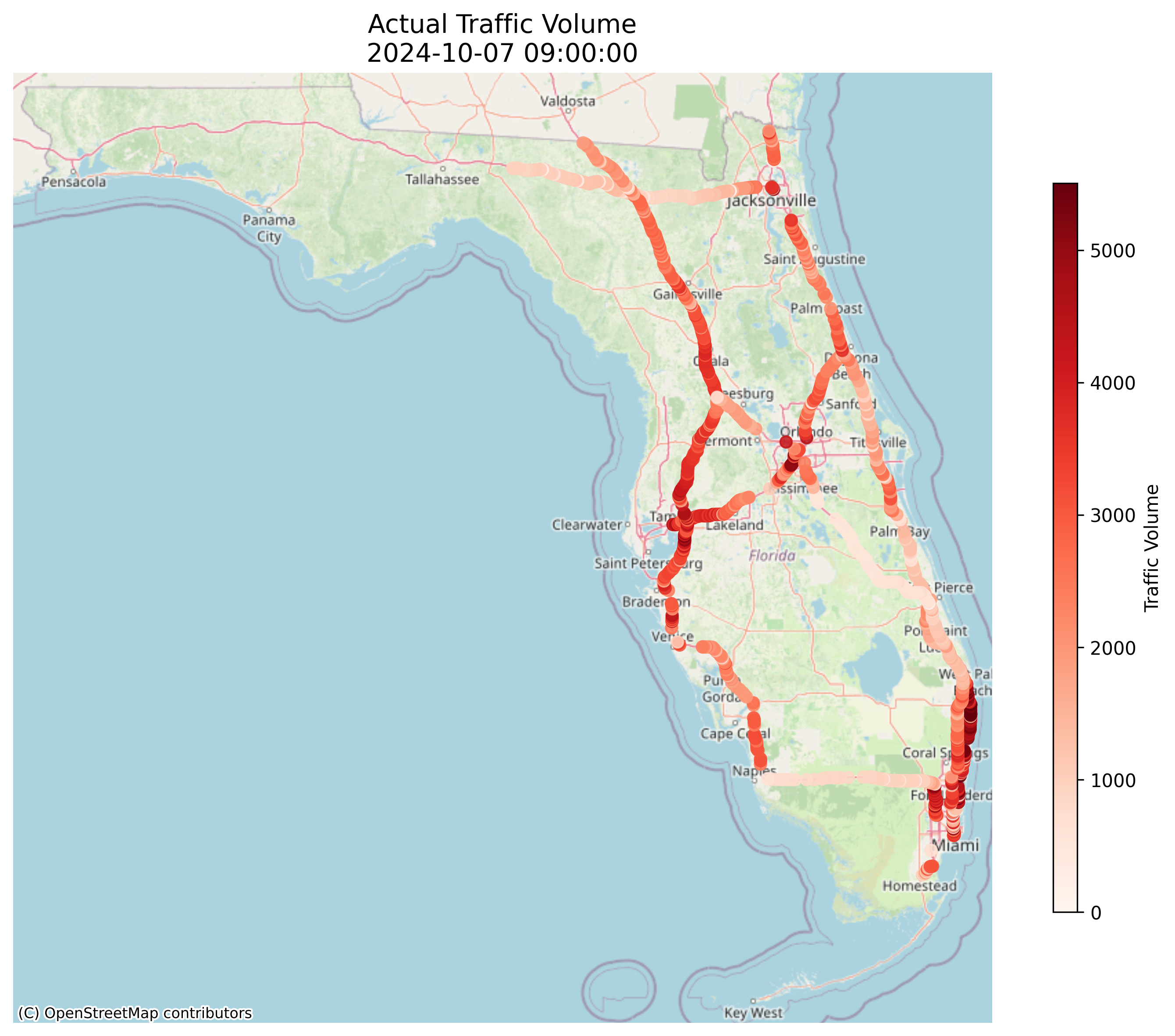}
        \caption{2024-10-07 09:00:00}
    \end{subfigure}
    \hspace{0.02\linewidth}
    \begin{subfigure}[b]{0.3\linewidth}
        \centering
        \includegraphics[width=\linewidth]{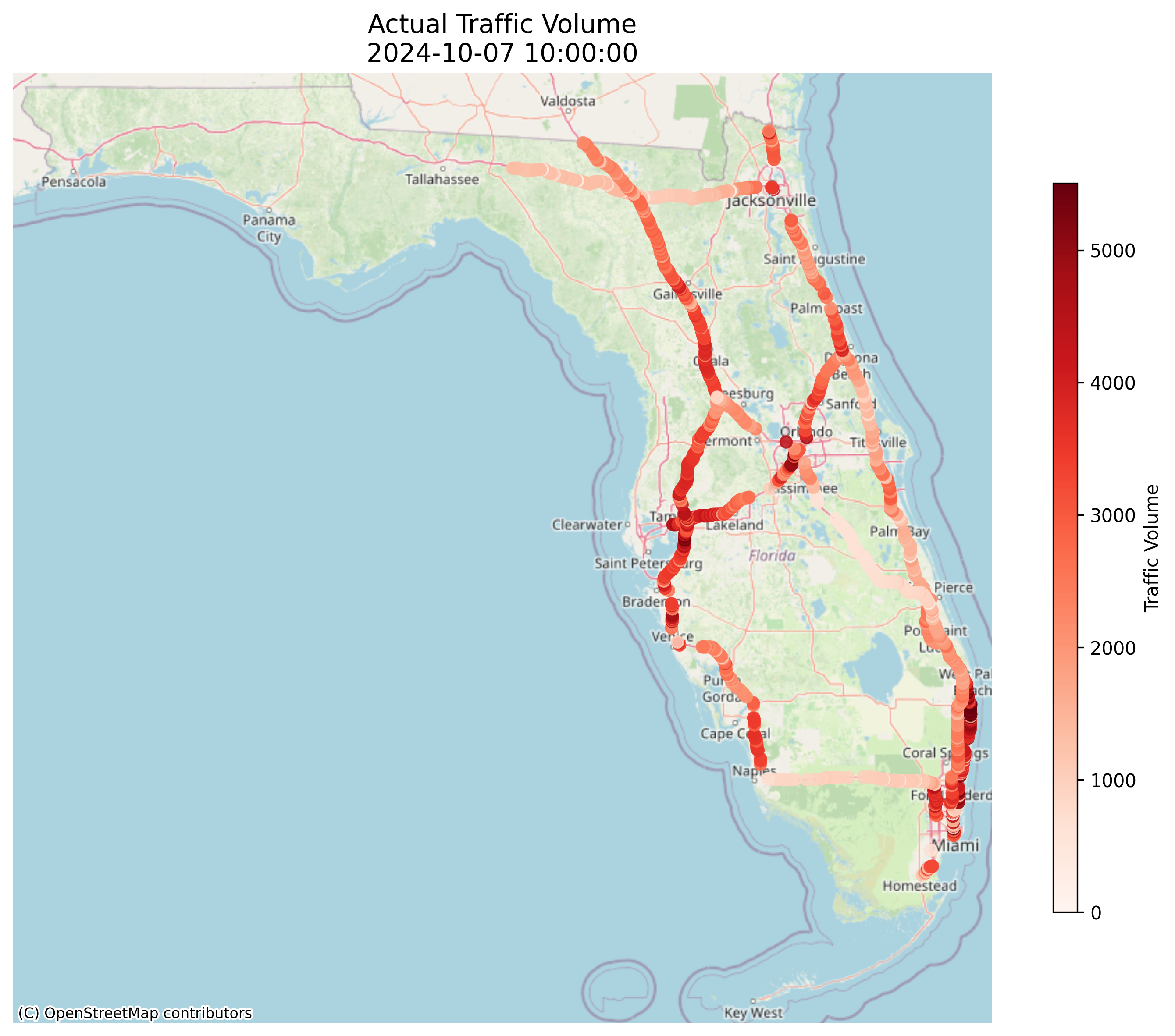}
        \caption{2024-10-07 10:00:00}
    \end{subfigure}
    \hspace{0.02\linewidth}
    \begin{subfigure}[b]{0.3\linewidth}
        \centering
        \includegraphics[width=\linewidth]{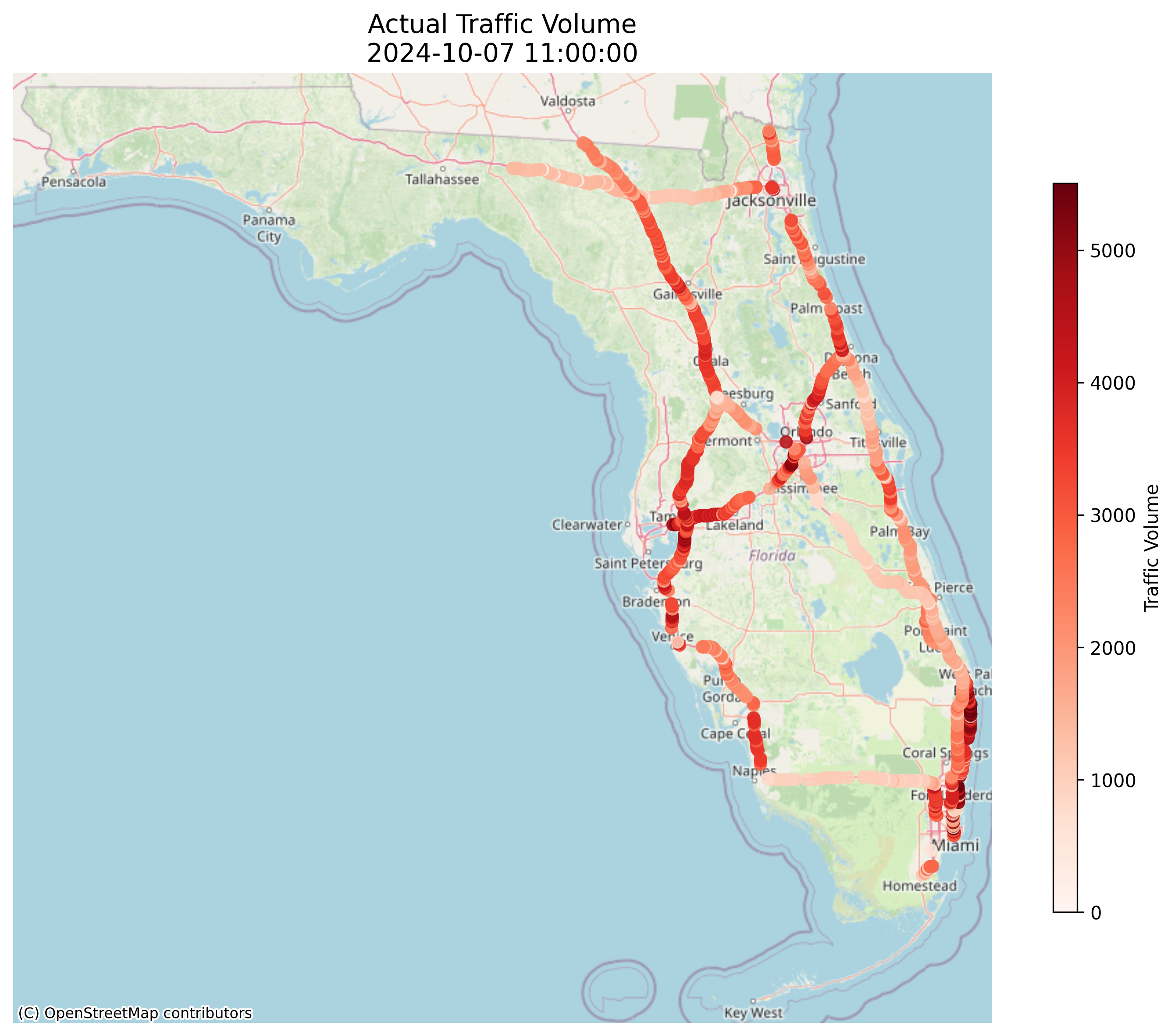}
        \caption{2024-10-07 11:00:00}
    \end{subfigure}

    \par\medskip
    (a) Congestion Propagation of Actual Traffic Flow

    \vspace{0.5cm}

    \begin{subfigure}[b]{0.3\linewidth}
        \centering
        \includegraphics[width=\linewidth]{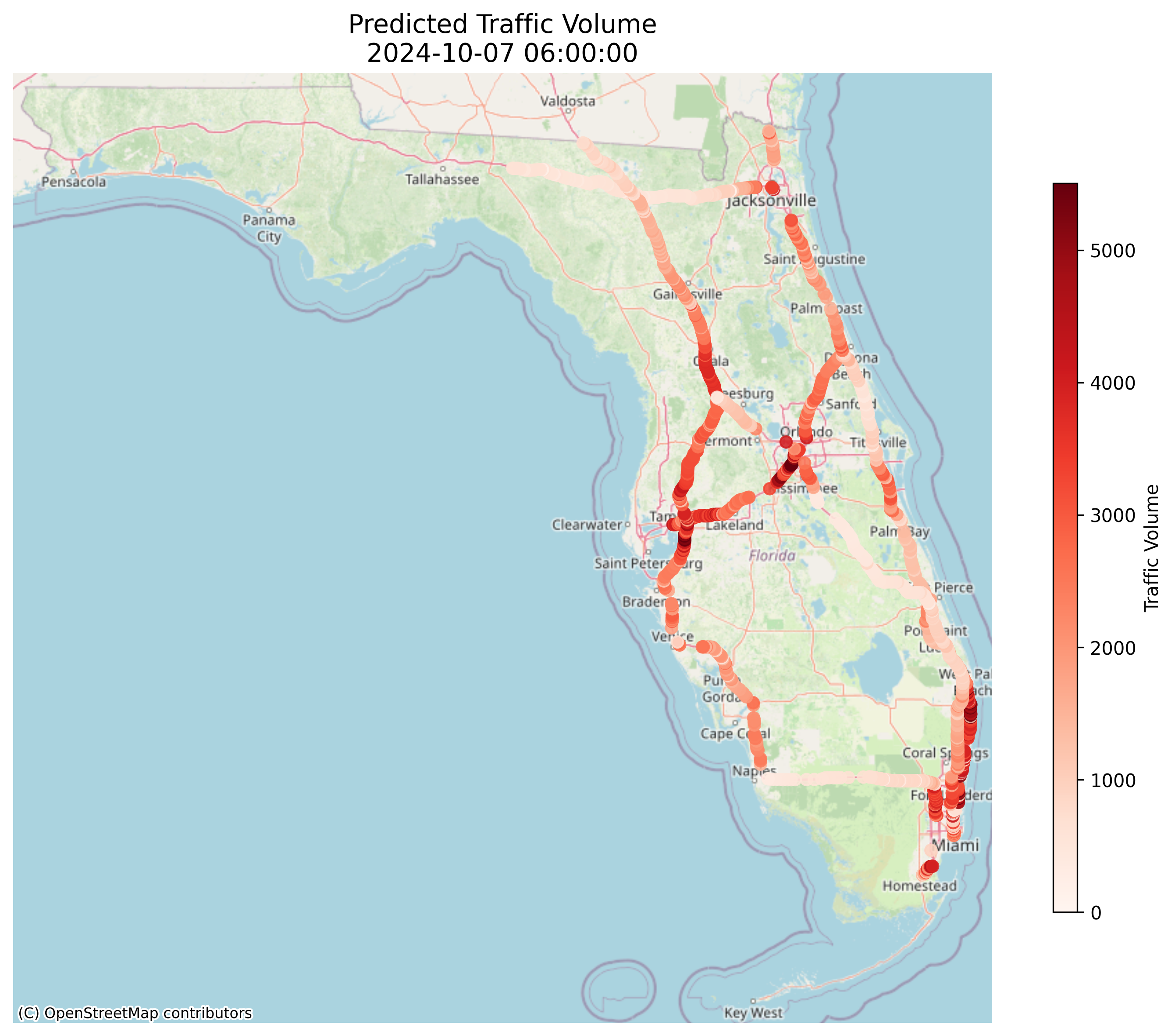}
        \caption{2024-10-07 06:00:00}
    \end{subfigure}
    \hspace{0.02\linewidth}
    \begin{subfigure}[b]{0.3\linewidth}
        \centering
        \includegraphics[width=\linewidth]{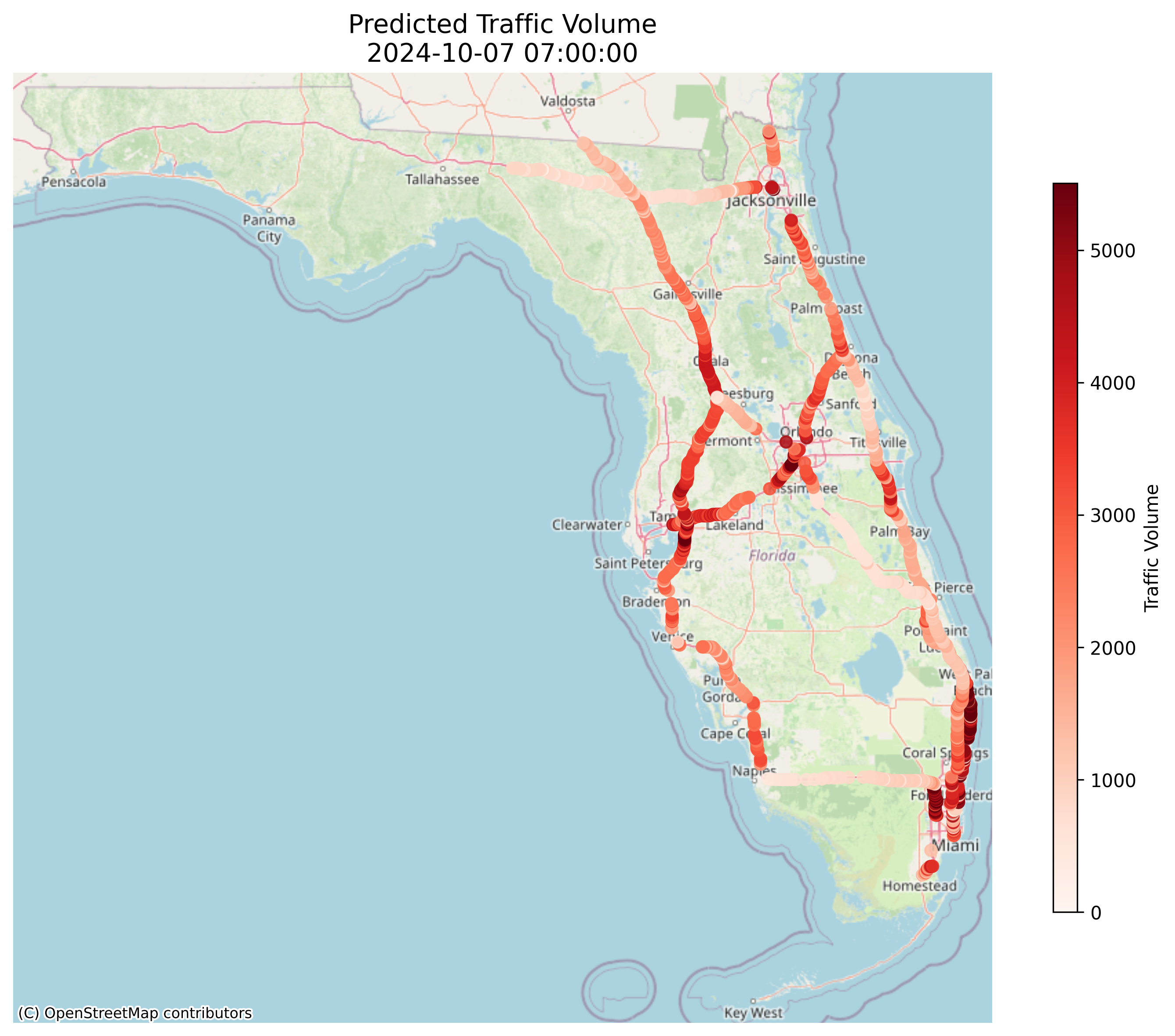}
        \caption{2024-10-07 07:00:00}
    \end{subfigure}
    \hspace{0.02\linewidth}
    \begin{subfigure}[b]{0.3\linewidth}
        \centering
        \includegraphics[width=\linewidth]{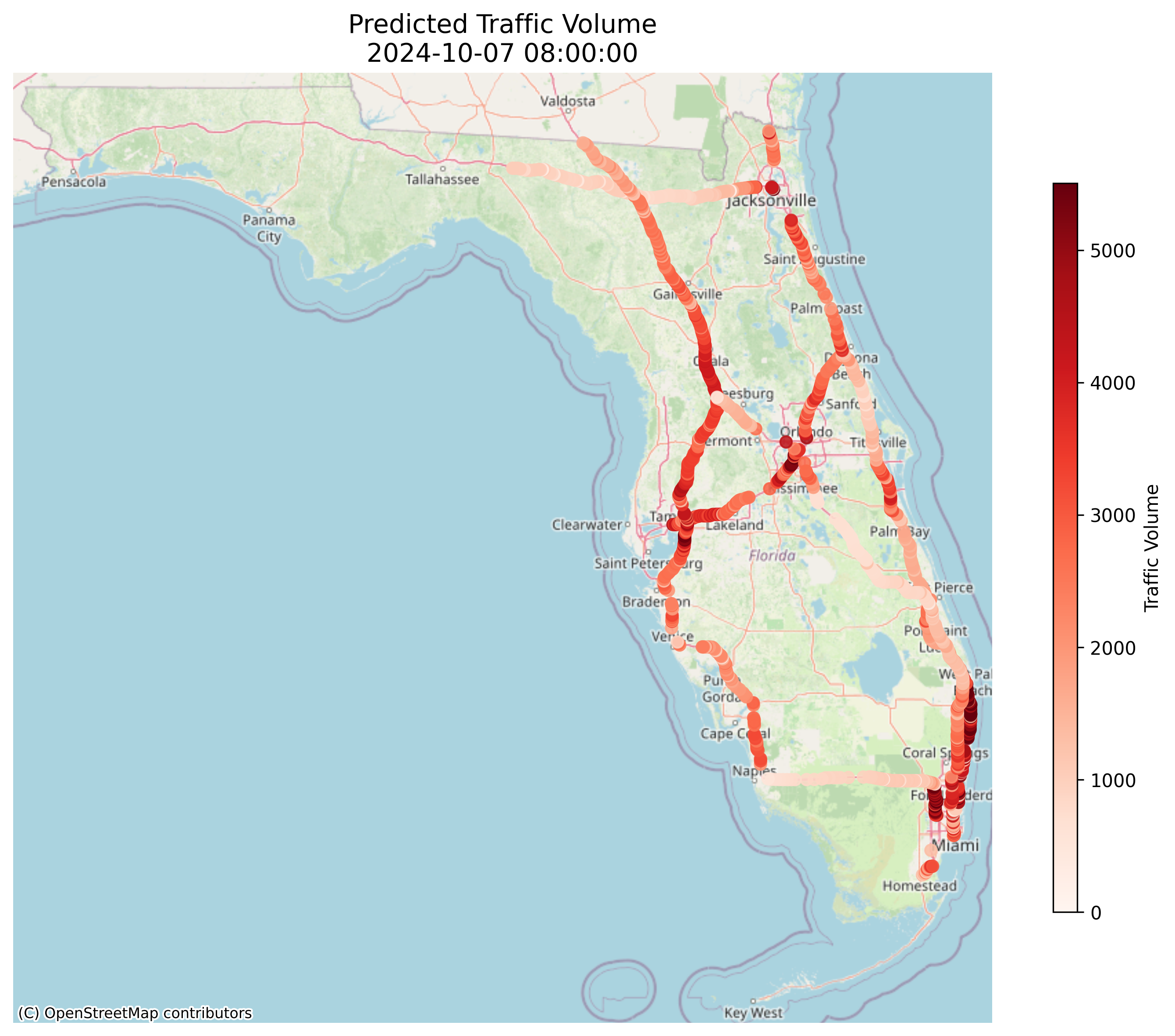}
        \caption{2024-10-07 08:00:00}
    \end{subfigure}

    \vspace{0.3cm}

    \begin{subfigure}[b]{0.3\linewidth}
        \centering
        \includegraphics[width=\linewidth]{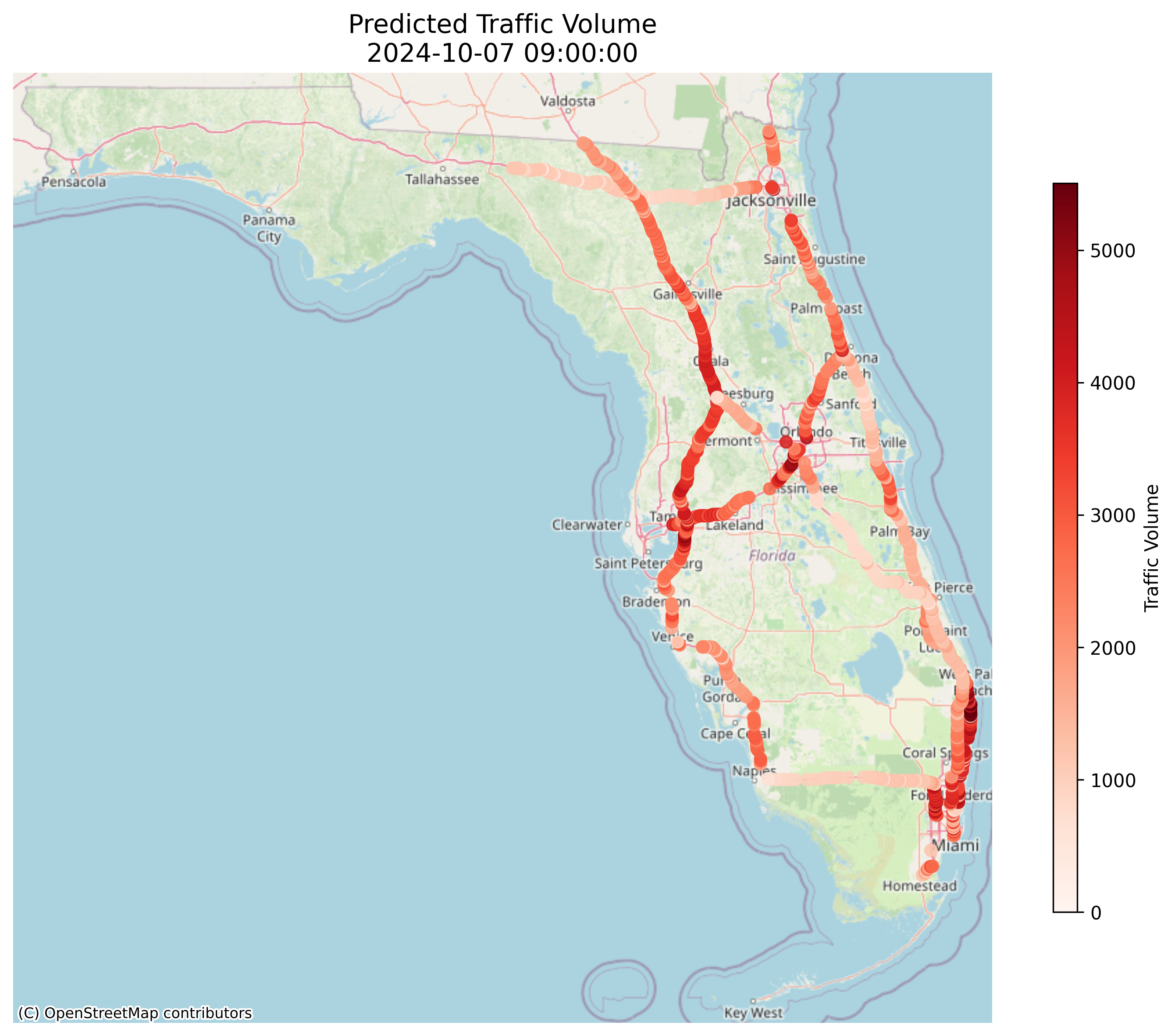}
        \caption{2024-10-07 09:00:00}
    \end{subfigure}
    \hspace{0.02\linewidth}
    \begin{subfigure}[b]{0.3\linewidth}
        \centering
        \includegraphics[width=\linewidth]{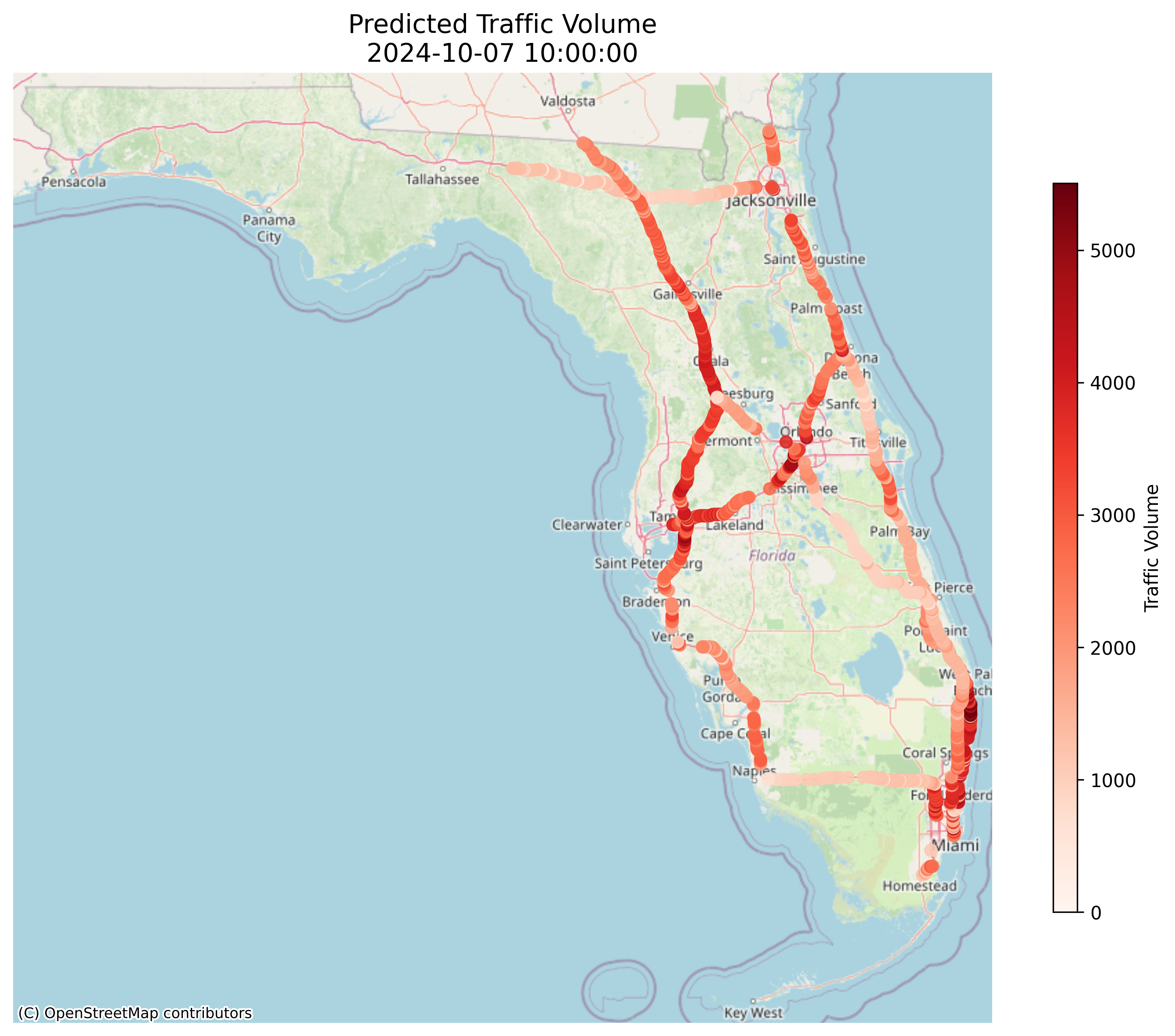}
        \caption{2024-10-07 10:00:00}
    \end{subfigure}
    \hspace{0.02\linewidth}
    \begin{subfigure}[b]{0.3\linewidth}
        \centering
        \includegraphics[width=\linewidth]{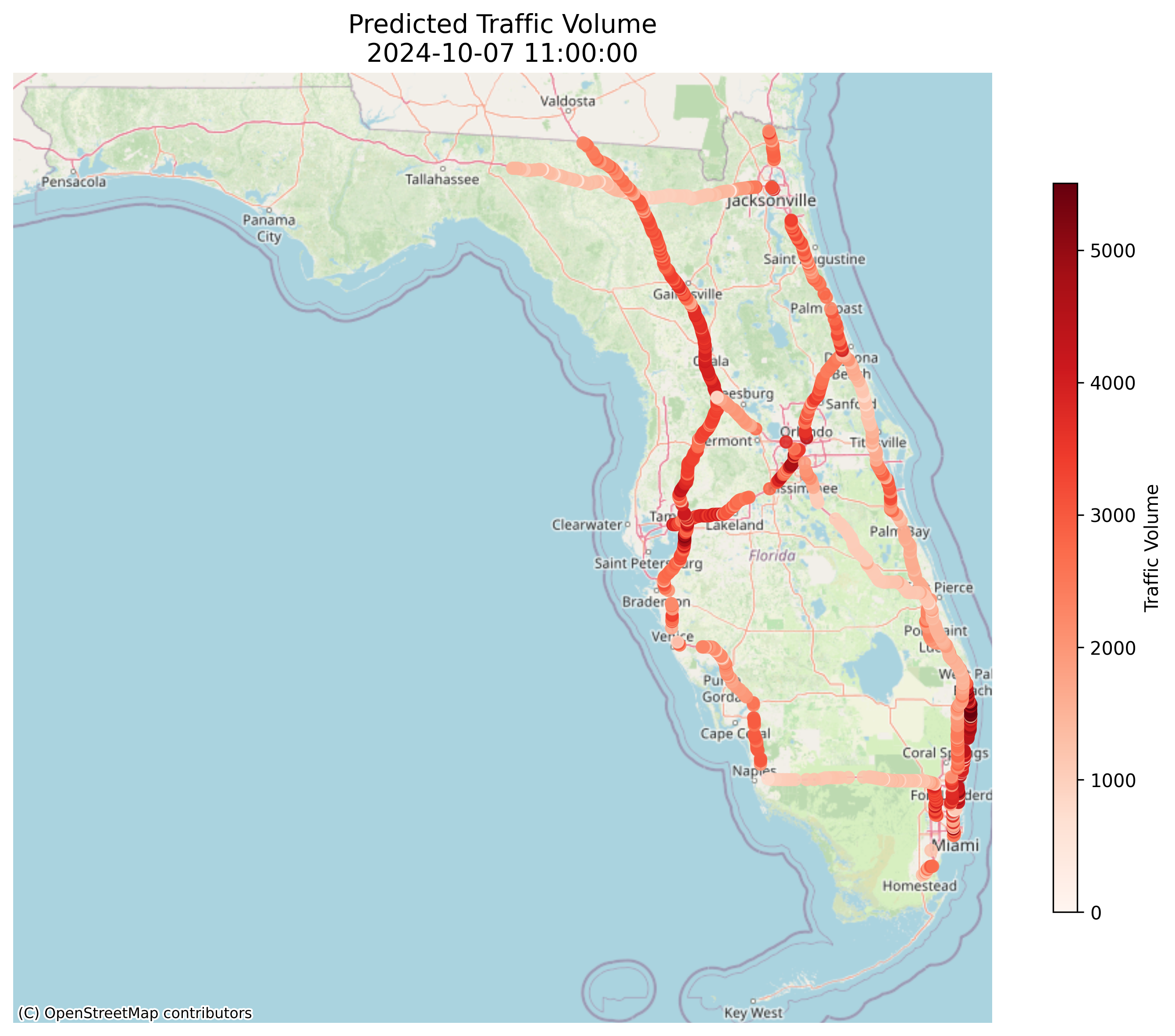}
        \caption{2024-10-07 11:00:00}
    \end{subfigure}

    \par\medskip
    (b) Congestion Propagation of Predicted Traffic Flow

    \caption{Network-Wide Congestion Propagation of (a) Actual and (b) Predicted Traffic Flow during Milton}
    \label{fig:congestion_actual_vs_predicted}
\end{figure}

Figure~\ref{fig:congestion_actual_vs_predicted} compares the congestion propagation patterns of actual and predicted traffic flows over a 6-hour horizon. The alignment between actual and predicted flows indicates that the proposed model successfully captures spatial and temporal evolution of congestion during evacuation. The predicted congestion hotspots closely resemble the actual ones in both location and intensity. It demonstrates the model's effectiveness in learning and reproducing dynamic evacuation traffic patterns.

\begin{figure}[htbp]
    \centering
    \includegraphics[width=0.9\linewidth]{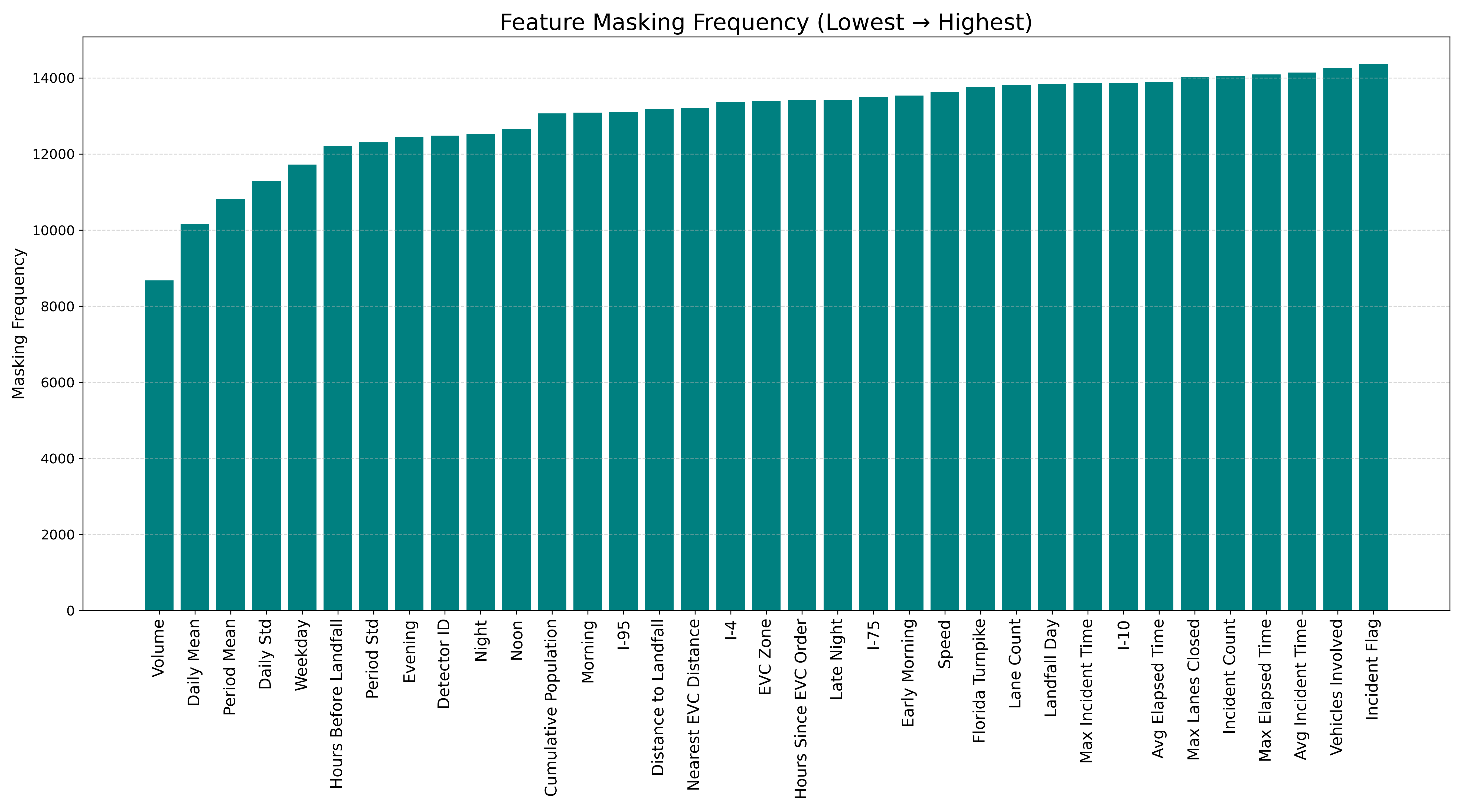}
    \caption{RL-IFSR: DDQN based feature masking frequency}
    \label{fig:fet_mask_fr}
\end{figure}

Figure~\ref{fig:fet_mask_fr} shows the frequency of each feature being masked by RL-based Intelligent Feature Selection and Ranking (RL-IFSR) module during training. A lower masking frequency implies a higher importance as the model tends to retain more informative features. From the figure, it is evident that volume is the most critical feature, being masked the least number of times throughout training. This finding aligns well with the prediction task, as the model aims to forecast future traffic volume, making past volume data highly relevant. Additionally, features representing previous volume related statistics such as Previous Daily Mean and Standard Deviation, Previous Period Mean and Standard Deviation—also appear among the least masked features. This suggests that the model considers not only immediate past volume but also its historical patterns and variability as highly informative for predicting future evacuation traffic flows. Following volume features, weekday, hours before landfall, time-of-day indicators (e.g., Evening, Night, Noon), and cumulative population were among the least masked features, indicating their strong relevance to evacuation traffic dynamics. On the other hand, incident-related variables, and specific road type indicators were masked more frequently, suggesting their lower importance for prediction.   

By dynamically identifying and masking less informative features during training through RL-IFSR module, the model is encouraged to focus on a compact subset of critical features that are most predictive across diverse scenarios. This selective exposure acts as an implicit regularization technique, reducing overfitting to idiosyncrasies in training data. Furthermore, the agent’s exploration of different feature subsets enables the model to perform reliably even in cases of missing or noisy data, thereby enhancing its robustness. As a result, the model learns to make accurate predictions under a wider variety of real-world scenarios, including unseen evacuation traffic patterns, varying population responses, or sensor dropouts.

\section{Ablation Study} \label{sec:ablation}
To evaluate the contributions of each component of RL-DMF framework, we conduct an ablation study by systematically removing specific modules. This analysis helps isolate the impact of multi-graph structure and reinforcement learning-based feature selection mechanism on overall model performance. We consider the following variants:

\begin{itemize}
    \item \textbf{RL-DGL (Distance-based)}: This variant removes the travel-time-based graph and relies solely on distance-based graph. It retains the reinforcement learning (RL) feature masking module. 

    \item \textbf{RL-DGL (travel-time-based)}: This configuration omits the distance-based graph and uses only travel-time-based dynamic graph to encode spatial dependencies. The RL-based feature selection module remains active. 

    \item \textbf{DMF (Fusion w/o RL)}: In this setting, the model retains both dynamic graphs and multi-graph fusion mechanism but disables the reinforcement learning component for intelligent feature masking. All features are used without masking. This comparison highlights the effect of RL-based Intelligent Feature Selection and Ranking (RL-IFSR) mechanism on model performance.
\end{itemize}

\begin{table}[htbp]
\centering
\caption{Ablation Study Results for RL-DMF Framework for 6 hourly prediction horizons}
\label{tab:ablation_study}
{\scriptsize (\textit{Minimum flow = 6, Maximum flow = 10889, Mean flow = 1838, Median flow = 1503})}
{\small\begin{tabularx}{\linewidth}{l XXXXX}
\toprule
\textbf{Model} & \textbf{Horizon} & \textbf{RMSE} & \textbf{MAE} & \textbf{MAPE} & \textbf{R\textsuperscript{2}} \\
\midrule

\multirow{7}{*}{\parbox{3cm}{\centering RL-DGL\\(Distance-based)}}
  & 1-hour & 293.1 & 194.3 & 17.8 & 0.95 \\
  & 2-hour & 389.8 & 255.2 & 20.7 & 0.92 \\
  & 3-hour & 444.7 & 293.5 & 24.1 & 0.89 \\
  & 4-hour & 488.8 & 326.5 & 26.3 & 0.87 \\
  & 5-hour & 524.1 & 353.9 & 30.2 & 0.85 \\
  & 6-hour & 553.4 & 374.7 & 31.9 & 0.83 \\
  & \textbf{Overall} & \textbf{457.4} & \textbf{299.7} & \textbf{25.2} & \textbf{0.89} \\
\midrule

\multirow{7}{*}{\parbox{3cm}{\centering RL-DGL\\(Travel-time-based)}}
  & 1-hour & 303.9 & 195.6 & 17.2 & 0.95 \\
  & 2-hour & 399.7 & 258.4 & 20.4 & 0.92 \\
  & 3-hour & 452.8 & 297.2 & 23.5 & 0.89 \\
  & 4-hour & 483.1 & 320.0 & 25.7 & 0.87 \\
  & 5-hour & 508.1 & 340.1 & 29.5 & 0.86 \\
  & 6-hour & 532.9 & 360.6 & 31.5 & 0.84 \\
  & \textbf{Overall} & \textbf{453.3} & \textbf{295.3} & \textbf{24.6} & \textbf{0.89} \\
\midrule

\multirow{7}{*}{DMF (Fusion w/o RL)}
  & 1-hour & 285.3 & 187.4 & 17.6 & 0.96 \\
  & 2-hour & 385.3 & 255.4 & 21.3 & 0.92 \\
  & 3-hour & 445.3 & 298.5 & 24.3 & 0.89 \\
  & 4-hour & 484.0 & 326.5 & 26.6 & 0.87 \\
  & 5-hour & 511.2 & 347.2 & 30.0 & 0.86 \\
  & 6-hour & 529.7 & 362.8 & 33.0 & 0.84 \\
  & \textbf{Overall} & \textbf{448.0} & \textbf{296.3} & \textbf{25.5} & \textbf{0.89} \\
\midrule

\multirow{7}{*}{\parbox{3cm}{\centering \textbf{RL-DMF}\\\textbf{(Full Model)}}}
  & 1-hour & 293.9 & 189.5 & 17.9 & 0.95 \\
  & 2-hour & 380.0 & 248.4 & 20.8 & 0.92 \\
  & 3-hour & 430.6 & 285.1 & 23.9 & 0.90 \\
  & 4-hour & 455.0 & 304.9 & 27.2 & 0.89 \\
  & 5-hour & 471.4 & 320.3 & 29.7 & 0.88 \\
  & 6-hour & 495.3 & 338.2 & 31.7 & 0.86 \\
  & \textbf{Overall} & \textbf{426.4} & \textbf{281.1} & \textbf{25.2} & \textbf{0.90} \\
\bottomrule
\end{tabularx}}
\end{table}

Table~\ref{tab:ablation_study} presents the results of ablation study. The first two variants, RL-DGL (Distance-based) and RL-DGL (travel-time-based), perform reasonably well, achieving RMSE values of 457.4 and 453.3, respectively, and comparable $R^2$ scores of 0.89. This suggests that each graph type individually captures meaningful spatial structure for evacuation traffic. When both graphs are fused without reinforcement learning (DMF), the performance slightly improves (RMSE = 448.0, $R^2$ = 0.89) compared to previous variants. Additionally, DMF achieves better predictive accuracy than using either the distance-based or travel-time-based dynamic GCN-LSTM models individually (as reported in Table~\ref{tab:baseline_comparison}). However, the full RL-DMF model, which integrates both graph modalities and leverages reinforcement learning for intelligent feature selection, achieves the best performance with an RMSE of 426.4 and an $R^2$ of 0.90. These results demonstrate the importance of both multi-graph fusion and RL-based feature selection in enhancing prediction accuracy.

\section{Policy Implications}
The proposed RL-DMF framework has the potential to improve evacuation traffic management and policies. Effective evacuation management requires a complete understanding of roadway traffic conditions before evacuation orders are issued \cite{Staes2021}. Based on the traffic conditions, transportation agencies can undertake traffic control strategies such as emergency shoulder use \cite{fdot_esu}, contraflow operations \cite{wolshon2005review2}, signal timing adjustments, route control \cite{FHWA2006, Houston2006} etc for an effective evacuation operation. Accurate forecasting of evacuation traffic can enable authorities to make proactive decisions about appropriate strategies \cite{wolshon2005review1, FHWA2022}. The proposed RL-DMF model can provide dynamic network-wide forecasts of traffic conditions in real-time. By predicting future traffic volumes across the whole network for next 1-hour to 6-hour horizons, transportation agencies can identify potential congestion hotspots in advance (as shown in Figure~\ref{fig:congestion_actual_vs_predicted}). The model can also capture the dynamic congestion propagation pattern of future traffic, based on which authorities can implement dynamic traffic control strategies to balance network loads. Such control can prevent the formation of severe bottlenecks, particularly along major evacuation corridors and improve overall evacuation efficiency and safety.

Additionally, interventions such as emergency shoulder use and contraflow require substantial lead time, staffing, and inter-agency coordination, and are most effective when activated under the right traffic conditions. Emergency management centers can use the predictive outputs from the proposed RL-DMF model to determine optimal timing and spatial allocation of these interventions. Moreover, the framework will enable data-driven prioritization of resources—such as positioning of law enforcement units, fuel supply points, and roadside assistance vehicles—based on predicted congestion zones. These capabilities can improve the overall situational awareness and public safety during extreme events \cite{wolshon2005review1}.

The reinforcement learning–guided feature selection also improve the trustworthiness of the proposed data-driven model for evacuation traffic prediction. Features that are identified as important by the RL-DMF model may help to improve the evacuation management efforts. Policymakers can leverage these explainable insights to understand which traffic, evacuation, or incident-related factors most strongly associated evacuation dynamics. This knowledge can support better evacuation preparation and management. For example, prior research demonstrated that weekday evacuation traffic may interact with regular peak-period commuting flows \cite{Dow2002,Cheng2013}. To be specific, work and school schedules on weekdays may delay or concentrate the evacuation departures \cite{Sadri2013,McCaffrey2018}. Workplace obligations on weekdays also influence when households can evacuate \cite{Lindell2019}. The RL-DMF model also find weekday indicator as an important predictor for evacuation traffic (see Figure~\ref{fig:fet_mask_fr}). It suggests that evacuation operations conducted on weekdays require additional planning and coordination. Agencies can use this information to issue evacuation orders in such a way to avoid peak commuting periods. They can coordinate institutional closures (schools, universities, major employers) and prepare for higher traffic in urban areas and near employment hubs.

Additionally, the RL-DMF framework is generalized across multiple hurricane events and regions. Evacuation behavior varies widely across hurricanes due to differences in hurricane track, intensity, timing, evacuation order dissemination, and levels of public response \cite{Lindell2019,Dow2002}.  Regional differences in roadway networks, population distributions, and institutional practices further complicate the transfer of evacuation traffic management practices across hurricanes. As network and demand patterns are not consistent from one event to another, emergency managers require generalized predictive tools that can perform reliably under a broad range of hurricane conditions \cite{MurrayTuite2013}. As the RL-DMF model is network-wide and trained across multiple hurricanes, policies informed by this model can be applied consistently across counties and regions, rather than being limited to a single corridor or a single evacuation event. This enables agencies to develop unified and standardized evacuation procedures, harmonize clearance time estimates across jurisdictions. 

Importantly, the RL-DMF model is designed to predict evacuation traffic for any future hurricane of Florida without any model retraining or fine-tuning. In practice, this means that as soon as a future hurricane forms and its projected path and timing become available, emergency managers can generate accurate, statewide real-time traffic forecasts and take interventions in highly congested regions. Such readiness is essential especially during rapidly intensifying hurricanes, in which the evacuation preparation time is very limited \cite{Bhatia2019_RI,Lipiec2024_RI}. By eliminating the need for retraining or fine-tuning, the proposed framework will enable faster and more reliable real-time evacuation traffic management.

To effectively deploy a predictive model for evacuation traffic management, transportation agencies may need to update their operational policies under emergency evacuations. These updates include real-time data sharing, improved coordination among agencies, and the responsible use of AI-based decision support tools. In particular, agencies will need to establish protocols for exchanging traffic detector data, weather updates, shelter availability, and evacuation order information across local, county, and state jurisdictions. Real-time integration of these datasets is essential for ensuring that the predictive model reflects evolving conditions and produces accurate forecasts. Effective deployment may also require coordination between transportation agencies, law enforcement, and emergency management agencies so that predictive insights translate into unified operational responses such as synchronized activation of any traffic management intervention (e.g., emergency shoulder use, contraflow), consistent public messaging, and coordinated routing guidance across neighboring counties or districts.

\section{Conclusions} \label{sec:conclusions}
This study introduces a novel Reinforcement Learning-guided Dynamic Multigraph Fusion (RL-DMF) framework for network-wide evacuation traffic prediction. Compared to state-of-the-art baseline models, the developed framework enhances the predictive accuracy, model interpretability, and ensures robust performance under complex, rapidly evolving evacuation traffic conditions. Ablation experiments demonstrate the significance of integrating DMF and RL-IFSR module for accurate prediction. The proposed model has substantial potential of strengthening hurricane evacuation management through predictive accuracy and interpretable decision support. Accurate forecasts allow emergency managers to detect congested areas in advance and deploy strategies to mitigate the congestion, improving the evacuation efficiency and safety.

While the RL-DMF framework shows strong predictive performance and generalizability, there are some limitations. The model's learning process is primarily based on historical traffic patterns, which limits its ability to adapt to abrupt external changes such as real-time weather conditions, sudden road closures, or emergency traffic control interventions. Additionally, although reinforcement learning effectively ranks feature importance, it assumes a shared action space across all nodes. Future research could explore node-specific or region-specific feature importance to further improve performance. Furthermore, the current model has only been evaluated for evacuation traffic prediction within the state of Florida. To extend its applicability to other regions across, domain adaptation techniques may be required to address differences in regional traffic patterns, infrastructure, and evacuation behaviors. Additionally, adoption of the proposed framework may require policy updates emphasizing real-time data sharing and inter-agency coordination.

\section*{Funding}
The study was partially supported by U.S. National Science Foundation (NSF) EAGER Grant No. 2122135. However, the authors are solely responsible for the findings presented here.

\section*{Author Contributions Statement (CRediT)}
\textbf{M. Rafi:} Conceptualization, Data curation, Formal analysis, Investigation, Methodology, Validation, Visualization, Writing – original draft.  
\textbf{S. Hasan:} Conceptualization, Methodology, Funding acquisition, Resources, Supervision, Project administration, Validation, Writing – review and editing.  
Both authors reviewed and approved the final manuscript.

\section*{Declaration of Competing Interest}
The authors declare that they have no competing financial interests or personal relationships that could have influenced the work reported in this study.

\section*{Data Availability Statement}
A request has been sent to the raw data owner (Florida Department of Transportation) for publishing the data associated with this study. The data will be made available once the request is approved.

\section*{Declaration of generative AI}
During the preparation of this manuscript, the authors used ChatGPT for assistance in language refinement and LaTeX formatting. After using this tool/service, the authors reviewed and edited the content as needed and take full responsibility for the content of the publication.

\bibliography{all}
\end{document}